\documentclass{article} % For LaTeX2e
\usepackage{iclr2022_conference,times}

\usepackage{amsmath,amsfonts,bm}

\def\eqref#1{equation~\ref{#1}}
\def\1{\bm{1}}

\DeclareMathAlphabet{\mathsfit}{\encodingdefault}{\sfdefault}{m}{sl}
\SetMathAlphabet{\mathsfit}{bold}{\encodingdefault}{\sfdefault}{bx}{n}

\usepackage{url}
\usepackage{algorithm}
\usepackage{algorithmic}
\usepackage{color}
\usepackage{amsthm}
\usepackage{makecell}
\usepackage{wrapfig}
\usepackage{graphicx}
\usepackage{float}
\usepackage{subfigure}
\usepackage{verbatim}
\usepackage[colorlinks,linkcolor=black,citecolor=blue]{hyperref}
\usepackage{multirow}
\newtheorem{assumption}{Assumption}
\newtheorem{lemma}{Lemma}
\newtheorem{theorem}{Theorem}

\usepackage{wrapfig}
\usepackage{amsmath}

\usepackage{graphicx} %插入图片的宏包
\usepackage{float} %设置图片浮动位置的宏包
\usepackage{subfigure}
\usepackage{enumitem}
\title{Concurrent Adversarial Learning for Large-Batch Training}

\author{Yong Liu\textsuperscript{1}, Xiangning Chen\textsuperscript{2}, Minhao Cheng\textsuperscript{3}, Cho-Jui Hsieh\textsuperscript{2}, Yang You\textsuperscript{1} \\
\textsuperscript{1}Department of Computer Science, National University of Singapore \\
\textsuperscript{2}Department of Computer Science,  University of California, Los Angeles \\
\textsuperscript{3}Department of Computer Science and Engineering, Hong Kong University of Science and Technology \\
\texttt{\{liuyong, youy\}@comp.nus.edu.sg}, \texttt{\{xiangning, chohsieh\}@cs.ucla.edu}, \\
\texttt{minhaocheng@ust.hk} \\
}

\begin{document}

\maketitle

\begin{abstract}

Large-batch training has become a commonly used technique when training neural networks with a large number of GPU/TPU processors. As batch size increases, stochastic optimizers tend to converge to sharp local minima, leading to degraded test performance. Current methods usually use extensive data augmentation to increase the batch size, but we found the performance gain with data augmentation decreases as batch size increases, and data augmentation will become insufficient after certain point. In this paper, we propose to use adversarial learning to increase the batch size in large-batch training. Despite being a natural choice for smoothing the decision surface and biasing towards a flat region, adversarial learning has not been successfully applied in large-batch training since it requires at least two sequential gradient computations at each step, which will at least double the running time compared with vanilla training even with a large number of processors. To overcome this issue, we propose a novel Concurrent Adversarial Learning (ConAdv) method that decouple the sequential gradient computations in adversarial learning by utilizing staled parameters. Experimental results demonstrate that ConAdv can successfully  increase the batch size on ResNet-50 training on ImageNet while maintaining high accuracy. In particular, we show ConAdv along can achieve 75.3\% top-1 accuracy on ImageNet ResNet-50 training with 96K batch size, and the accuracy can be further improved to 76.2\% when combining ConAdv with data augmentation. This is the first work successfully scales ResNet-50 training batch size to 96K. 

\end{abstract}

\section{Introduction}
\label{section1}

With larger datasets and bigger models proposed, training neural networks has become quite time-consuming. For instance, training BERT \citep{DevlinCLT19} takes 3 days on 16 v3 TPUs. GPT-2 \citep{radford2019language} contains 1,542M parameters and requires 168 hours of training on 16 v3 TPU chips. This also leads to the developments of high performance computing clusters. For example, Google and NVIDIA build high performance clusters with thousands of TPU or GPU chips. How to fully utilize those computing resources for machine learning training thus becomes an important problem. 

Data parallelism is a commonly used technique for distributed neural network training, where each processor computes the gradient of a local batch and the gradients across processors are aggregated at each iteration for a parameter update. Training with hundreds or thousands of processors with data parallelism is thus equivalent to running a stochastic gradient optimizer (e.g., SGD or Adam) with a very large batch size, also known as {\bf large batch training}. For example, Google and NVIDIA showed that by increasing the batch size to 64k on ImageNet, they can finish 90-epoch ResNet training within one minute~\citep{kumar2021exploring,mattson2019mlperf}. 

But why can't we infinitely increase the batch size as long as more  computing resources are available? Large batch training often faces two challenges. First, under a fixed number of training epochs, increasing the batch size implies reducing number of training iterations.  Even worse, it has been observed that large-batch training often converges to solutions with bad generalization performance (also known as sharp local minima), possibly due to the lack of inherent noise in each stochastic gradient update. Although this problem can be partially mitigated by using different optimizers such as LARS~\citep{you2017scaling} and LAMB~\citep{you2019large}, the limit of batch size still exists. For instance, Google utilizes several techniques, such as distributed batch normalization and Mixed-precision training, to further scale the traing of ResNet-50 on 4096 v3 TPU chips. However, it can just expand the batch size to 64k\citep{kumar2021exploring,ying2018image}.

Data augmentation has become an indispensable component of large-batch training pipeline. For instance, researchers at Facebook use augmentation to scale the training of ResNet-50 to 256 NVIDIA P100 GPUs with a batch size of 8k on ImageNet \citep{goyal2017accurate}. You et al. also use data augmentation to expand the batch size to 32k on 2048 KNL nodes~\citep{you2018imagenet}. 
% Google use augmentation to further scale the training of ResNet-50 on v3-4096 TPU chips with a batch size of 64k.
%xxx (cite previous papers who say augmentation is useful and what's the best batch size they can achieve, for example, facebook using strong data augmentation to reach xx batch size; I think Google also use augmentation to scale imagenet batch to 64K). } 
% This is plausible since data augmentation can implicitly penalize sharp regions \xc{citations? My experience it's not penalizing sharp regions} and improve the generalization of the solution. 
However, in this paper we find that when batch size is large enough (i.e., larger than 32k), the increased diversity in augmented data will also increase the difficulty of training and even have a negative impact on test accuracy. 

This motivates us to study the application of adversarial training in large-batch training. Adversarial training methods find a perturbation within a bounded set around each sample to train the model. Previous work finds the adversarial training would lead to a significant decrease in the curvature
of the loss surface and make the network more "linear" in the small batch size case, which could be used as a way to improve generalization
% , which intuitively also provides a way to \xc{delete} avoid sharp local minima and improve generalization
~\citep{xie2020adversarial,moosavi2019robustness}. However, adversarial training has not been used in large-batch training since it requires a series of sequential gradient computations within each update to find an adversarial example. 
%One has to first run a gradient-based attacker (e.g., FGSM~\cite{goodfellow2014explaining} or PGD~\cite{madry2017towards}) for $k$ steps to get $x^{*}$, and then compute another gradient with $x^{*}$ to update the weight. 
Even when conducting only $1$ gradient ascent for finding adversarial examples, adversarial training requires $2$ sequential gradient computations (one for adversarial example and one for weight update) that cannot be parallelized.  Therefore, even with infinite computing resources, adversarial training is at least 2 times slower than standard training and increasing the batch size cannot compensate for that. 

To resolve this issue and make adversarial training applicable for large-batch training, we propose a novel Concurrent Adversarial Learning (ConAdv) algorithm for large-batch training. We show that by allowing the computation of adversarial examples using staled weights, the two sequential gradient computations in adversarial training can be decoupled, leading to fully parallelized computations at each step. As a result, extra processors can be fully utilized to achieve the same iteration throughput as original SGD or Adam optimizers. Comprehensive experimental results on large-batch training demonstrate that ConAdv is a better choice than existing augmentations.

Our main contributions are listed below: 
\begin{itemize}[leftmargin=*]
%\begin{compactitem}
% \item We analysis the effect of data augmentation from the pwerspective of Batch Normalization in large-batch training. We find that traditional data augmentation cannot be used in the large-batch training due to decreasing the iterations during training. The main reason is that the statics of BatchNorm is underfitting. In this paper, we use auxiliary BatchNorm can solve it and improve the original performance. 

\item This is the first work showing adversarial learning can significantly increase the batch size limit of large-batch training without using data augmentation.
\item The proposed algorithm, ConvAdv, can successfully  decouple the two sequential gradient computations in adversarial training and make them parallelizable. This makes adversarial training achieve similar efficiency with standard stochastic optimizers when using sufficient computing resources. Furthermore, we empirically show that ConAdv achieves almost identical performance as the original adversarial training. We also provide a theoretical analysis on ConAdv. 
\item Comprehensive experimental studies demonstrate that the proposed method can push the limit of large batch training on various tasks. For ResNet-50 training on ImageNet, ConAdv alone achieves 75.3\% accuracy when using 96K batch size. Further, the accuracy will rise to 76.2\% when combined with data augmentation. This is the first method scaling ResNet-50 batch size to 96K with accuracy matching the MLPerf standard (75.9\%), while previous methods fail to scale beyond 64K batch size. 
%We propose a distributed adversarial learning algorithm to make the model converge to the flat minima and escape from sharp minima. In this paper, the batch size can be expanded to more than 96k for ResNet-50 on ImageNet and we can use more computing resource to improve the efficiency of deep learning.
%\item Adversarial learning can expand batch size but cause extra time cost due to collectinng adversarial examples. In this paper, We propose a novel concurrent adversarial learning algorithm to further solve it, which use the stale weights to collect the adversarial examples and in this way we can obtain the adversarial data before updating the model at each step. Finally, we provide its convergence analysis.
%\item we provide the convergence analysis of proposed ConAdv algorithm, which illustrates that ConAdv can converge to an first-order stationary point at a sublinear rate up to a precision of $(L_{\theta x}^{2} \eta \tau M)^{2}$.
\end{itemize}
%\end{compactitem}

\section{Background}

% In this section, we will introduce several important concepts of this paper, such as large-batch training, adversarial learning and MLPerf.

\label{section2}

\subsection{Large-Batch Training}
\label{section2.1}

Using data parallelism with SGD naturally leads to  large-batch training on distributed systems. However, it was shown that extremely large batch is difficult to converge and has a  generalization gap \citep{KeskarMNST17,hoffer2017train} . Therefore, related work start to carefully fine-tune the hyper-parameters to bridge the gap, such as learning rate, momentum \citep{you2018imagenet,goyal2017accurate,li2017scaling,shallue2018measuring} . Goyal et al. try to narrow the generalization gap with the heuristics of learning rate scaling \citep{goyal2017accurate}. However, there is still big room to increase the batch size. Several recent works try to use adaptive learning rate to reduce the fine-tuning of hyper-parameters and further scaling the batch size to larger value \citep{you2018imagenet,iandola2016firecaffe,codreanu2017scale,akiba2017extremely,jia2018highly,smith2017don,martens2015optimizing,devarakonda2017adabatch,osawa2018second,you2019large,yamazaki2019yet}. 
You et al. propose Layer-wise
Adaptive Rate Scaling (LARS) \citep{you2017scaling} for better optimization and scaling to the batch size of 32k without performance penalty on ImageNet. Ying et al. use LARS optimizer to train ResNet-50 on TPU Pods in 2.2 minutes. In addition, related work also try to bridge the gap from aspect of augmentation. Goyal et al. use data augmentation to further scale the training of ResNet-50 on ImageNet \citep{goyal2017accurate}. Yao et al. propose an adaptive batch size method based on Hessian information to gradually increase batch size during training and use vanilla adversarial training to regularize against the sharp minima \cite{yao2018large}. However, the process of adversarial training is time consuming and they just use the batch size of 16k in the second half of training process (the initial batch size is 256). 
How to further accelerate the training process based on adversarial training and reduce its computational burden is still an open problem.

\subsection{Adversarial Learning}
\label{section2.2}

Adversarial training has shown great success on improving the model robustness through collecting adversarial examples and injecting them into training data \cite{goodfellow2014explaining,papernot2016transferability,wang2019convergence}. \cite{madry2017towards} formulates it into a min-max optimization framework as follows: 
% Adversarial learning can improve the robustness through collecting adversarial examples and injecting them into training data \cite{papernot2016transferability,goodfellow2014explaining}. The key idea is to modify the training objective through calculating the perturbation on the input data and maxmimize the modified loss function:
\begin{equation}
    \mathop{\min}\limits_{\theta} \mathbb{E}_{(x_{i},y_{i})\sim\mathbb{D}}[\mathop{\max}\limits_{||\delta||_{p}\in\epsilon}\mathcal{L}(\theta_{t},x+\delta,y)], 
\end{equation}
where $\mathbb{D}=\{(x_{i}, y_{i})\}_{i=1}^{n}$ denotes training samples and $x_{i} \in \mathbb{R}^{d}$, $y_{i} \in \{1, ..., Z\}$, $\delta$ is the adversarial perturbation, $||\cdot||_{p}$ denotes some $\mathcal{L}_{p}$-norm distance metric, $\theta_{t}$ is the parameters of time $t$ and Z is the number of classes. Goodfellow et al. proposes FGSM to collect adversarial data \cite{goodfellow2014explaining}, which performs a one-step update along the gradient direction (the sign) of the loss function. %Kurakin et al uses multi-step FGSM to generate adversarial examples iteratively \cite{kurakin2016adversarial}.

Project Gradient Descent (PGD) algorithm \cite{madry2017towards} firstly carries out random initial search in the allowable range (spherical noise region) near the original input, and then iterates FGSM several times to generate adversarial examples. Recently, several papers \cite{shafahi2019adversarial,wong2020fast, andriushchenko2020understanding} aim to improve the computation overhead brought by adversarial training. Specifically, FreeAdv \cite{shafahi2019adversarial} tries to update both weight parameter $\theta$ and adversarial example $x$ at the same time by exploiting the correlation between the gradient to the input and to the model weights. 
Similar to Free-adv, Zhang et al.~\cite{zhang2019you} further restrict most of the forward and back propagation within the first layer to speedup computation. 
FastAdv \cite{wong2020fast} finds the overhead could be further reduced by using single-step FGSM with random initialization. While these work aim to improve the efficiency of adversarial training, they still require at least two sequential gradient computations for every step. Our concurrent framework could decouple the two sequential gradient computation to further boost the efficiently, which is more suitable for large-batch training.  Recently, several works~\cite{xie2020adversarial, cheng2021adversarial, chen2021robust} show that the adversarial example can serve as an augmentation to benefit the clean accuracy in the small batch size setting. However, whether adversarial training can improve the performance of large-batch training is still an open problem.
%how to use adversarial training to achieve is still an unsolved problem.

\subsection{MLPerf}
\label{section2.3}

MLPerf \cite{mattson2019mlperf} is an industry-standard performance benchmark for machine learning, which aims to fairly evaluate system performance. Currently, it includes  several representative tasks from major ML areas, such as vision, language, recommendation. In this paper, we use ResNet-50 \cite{He_2016_CVPR} as our baseline model and the convergence baseline is 75.9\% accuracy on ImageNet.

\section{Proposed Algorithm}
\label{section 3}

In this section, we introduce our enlightening findings and the proposed algorithm. We first study the limitation of data augmentation in large-batch training. Then we discuss the bottleneck of adversarial training in distributed system and propose a novel Concurrent Adversarial Learning (ConAdv) method for large-batch training. 
%In addition, we provide the convergence analysis of proposed method. 
% Finally, we analysis how to make data augmentation and adversarial learning work together.

% data augmentation, in large-batch training, the properties are different from natural batch training.

% First, we try to use data augmentation to improve the performance of large-batch training.
% To achieve better performance, we try to use auxiliary BN to solve it.

% Furthermore, adversarial learning is a great way to improve the robustness and generalization of trained model.

% However, adversarial learning is cost since we need to calculate the gradients of batch data and backpropagation to collect the adversarial examples. Therefore, we try to propose a novel metric to determine whether the model need to collect the adversarial samples and only calculate the gradients when the decision is true.

\subsection{Does Data Augmentation Improve the Performance of Large-Batch Training?}
\label{section3.1}

\begin{wrapfigure} {r}{0.37\textwidth}
% \begin{minipage}[t]{.5\linewidth}
\centering
\includegraphics[width = 0.37\textwidth]{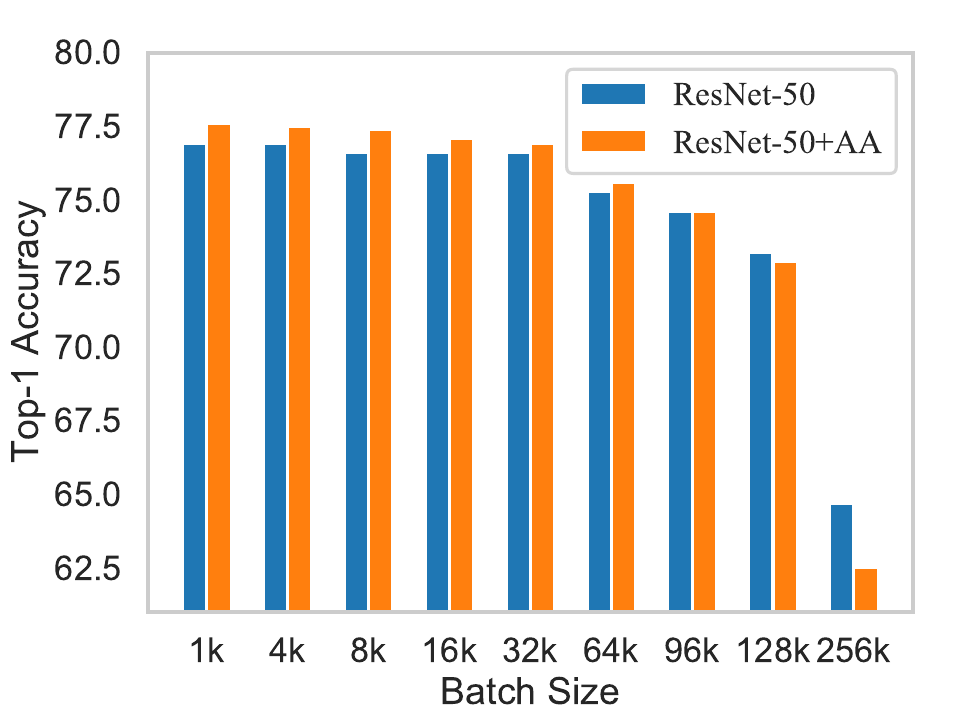}  
\caption{Augmentation Analysis}
\label{fig:aa_analysis}
% \end{minipage}
\end{wrapfigure}

Data augmentation can usually improve generalization of models and is a commonly used technique to improve the batch size limit in large-batch training. 
To formally study the effect of data augmentation in large-batch training, we train ResNet-50 using ImageNet~\cite{imagenet_cvpr09} by AutoAug (AA) \cite{Cubuk_2019_CVPR}. 
The results shown in Figure \ref{fig:aa_analysis} reveal that although AA helps improve generalization under batch size $\leq$ 64K, the performance gain decreases as batch size increases. Further, it could lead to negative effect when the batch size is large enough (e.g., 128K or 256K). For instance, the top-1 accuracy increase from $76.9\%$ to $77.5\%$ when using AA on 1k batch size. However, it decreases from 73.2\% to 72.9\% under data augmentation when the batch size is 128k and drops from 64.7\% to 62.5\% when the batch size is 256k. The main reason is that the augmented data increases the diversity of training data, which leads to slower convergence when using fewer training iterations. Recent work try to concat the original data and augmented data to jointly train the model and improve their accuracy \citep{berman2019multigrain}. However, we find that concating them will hurt the accuracy when batch size is large. Therefore, we just use the augmented data to train the model.
The above experimental results motivate us to explore a new method for large batch training.

\subsection{Adversarial Learning in the Distributed Setting}
\label{section3.2}
%缺少形式化的描述，感觉都是在讲白话

\begin{figure}

\centering
\includegraphics[width = 0.91\textwidth]{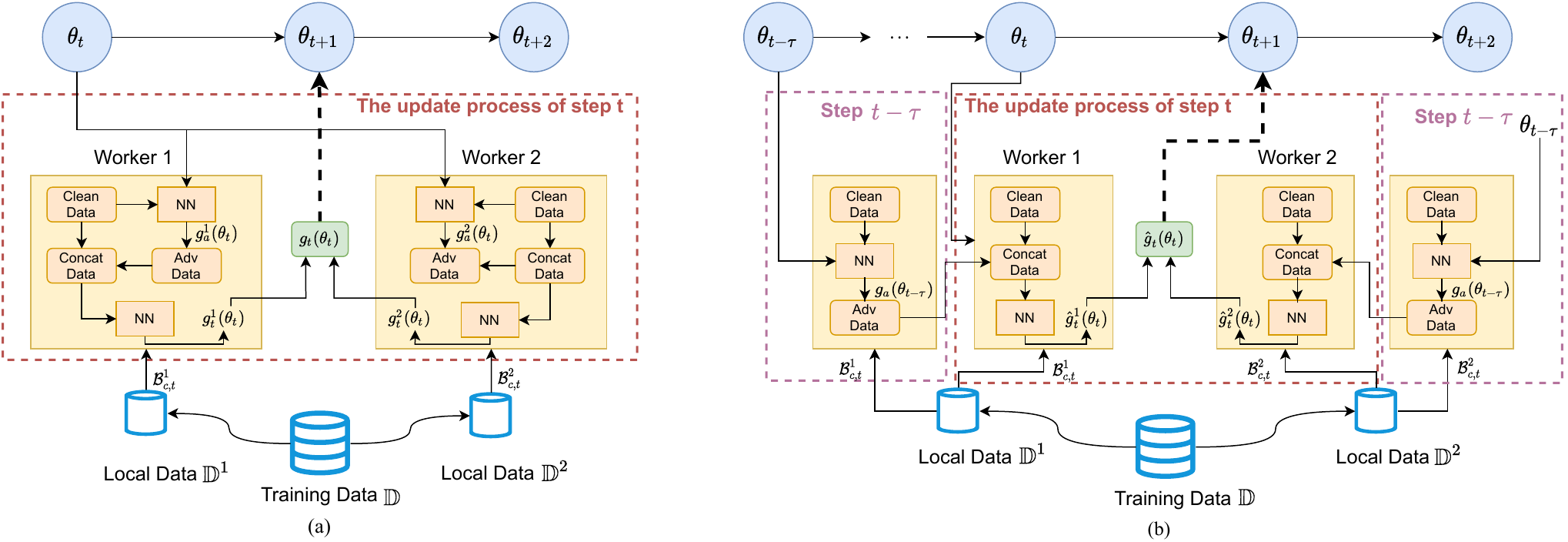}  
\caption{(a) Distributed Adversarial Learning (DisAdv), (b) Concurrent Adversarial Learning (ConAdv). To ease the understanding, we just show the system including two workers.
} 
\label{fig: structure}

\end{figure}

Adversarial learning can be viewed as a way to automatically conduct data augmentation. Instead of defining fixed rules to augment data, adversarial learning conducts gradient-based adversarial attacks to  find adversarial examples.
% ---points that near the original sample while encountering much larger loss. 
As a result, adversarial learning leads to smoother decision boundary~\cite{karimi2019characterizing,madry2017towards}, which often comes with flatter local minima~\cite{yao2018hessian}. Instead of solving the original empirical risk minimization problem, adversarial learning aims to solve a min-max objective that minimizes the loss under the worst case perturbation of samples within a small radius. In this paper, since our main goal is to improve clean accuracy instead of robustness, we consider the following training objective that includes loss on both natural samples and adversarial samples: 
\begin{equation}
\label{objective_function}
    \min_{\theta}  \mathbb{E}_{(x_{i},y_{i})\sim \mathbb{D}}[\mathcal{L}(\theta_{t}; x_{i}, y_{i}) + \max_{\|\delta\|_{p} \in \epsilon}\mathcal{L}(\theta_{t}; x_{i} + \delta, y_{i})],
\end{equation}
where $\mathcal{L}$ is the loss function 
%$K$ denotes the number of total workers 
and $\epsilon$ represents the value of perturbation. Although many previous work in adversarial training focus on improving the trade-off between accuracy and robustness~\cite{shafahi2019adversarial,wong2020fast}, recently Xie et al. show that using split BatchNorm for adversarial and clean data can improve the test performance on clean data~\cite{xie2020adversarial}. 
Therefore, we also adopt this split BatchNorm approach.

For Distributed Adversarial Learning (DisAdv), training data $\mathbb{D}$ is
partitioned into $N$ local dataset $\mathbb{D}^{k}$, and $\mathbb{D} = \cup_{k=1}^{k=K}\mathbb{D}^{k}$.
%, which is stored in each local machine (worker) of the cluster. 
For worker $k$, we firstly sample a mini-batch data (clean data) $\mathcal{B}_{t,c}^{k}$ from the local dataset $\mathbb{D}^{k}$ at each step $t$. After that, each worker downloads the weights $\theta_{t}$ from parameter sever and then uses $\theta_{t}$ to obtain the adversarial gradients $g_{a}^{k}(\theta_{t}) = \nabla_{x}\mathcal{L}(\theta_{t};x_{i},y_{i})$ on input example $x_{i} \in \mathcal{B}_{t,c}^{k}$. Noted that we just use the local loss $\mathbb{E}_{(x_{i},y_{i})\sim \mathbb{D}^{k}}\mathcal{L}(\theta_{t}; x_{i}, y_{i})$ to calculate the  adversarial gradient $g_{a}^{k}(\theta_{t})$ rather than the global loss $\mathbb{E}_{(x_{i},y_{i})\sim\mathbb{D}}\mathcal{L}(\theta_{t}; x_{i}, y_{i})$, 
since we aim to reduce the communication cost between workers. In addition, we use 1-step Project Gradient Descent (PGD) to calculate $\hat{x}_{i}^{*}(\theta_{t})=x_{i} + \alpha \cdot \nabla_{x}\mathcal{L}(\theta_{t};x_{i}, y_{i})$ to approximate the optimal adversarial example $x_{i}^{*}$. Therefore, we can collect the adversarial mini-batch  $\mathcal{B}_{a,t}^{k} = \{(\hat{x}_{i}^{*}(\theta_{t}), y_{i})\}$ and use both the clean example $(x_{i}, y_{i}) \in \mathcal{B}_{c,t}^{k}$ and adversarial example $(\hat{x}_{i}^{*}(\theta_{t}), y_{i}) \in \mathcal{B}_{a,t}^{k}$ to update the weights $\theta_{t}$. 
More specially, we use main BatchNorm to calculate the statics of clean data and  auxiliary BatchNorm to obtain the statics of adversarial data.
% Finally, we can update the weights of deep neural network through the total loss $\mathcal{L}(\theta_{t}, \mathcal{B}_{t}^{k})$.  In addition, to further speed up the training process, we set the iterations of adversarial attack is $1$. The pseudo code of proposed distributed adversarial learning algorithm is shown in Algorithm \ref{alg:FastAdv}. We will introduce ConAdv (Line 5) algorithm in the next section.

We show the workflow of adversarial learning on distributed systems (DisAdv) as Figure \ref{fig: structure}, and more importantly, we notice that it requires two sequential gradient computations at each step which is time-consuming and, thus, not suitable for large-batch training. 
Specifically, we firstly need to compute the gradient $g_{a}^{k}(\theta_{t})$ to collect adversarial example $\hat{x}^{*}$. After that, we use these examples to update the weights $\theta_{t}$, which computes the second gradient. 
In addition, the process of collecting adversarial example $\hat{x}_{i}^{*}$ and use $\hat{x}_{i}^{*}$ to update the model are tightly coupled, which means that each worker cannot calculate local loss $\mathbb{E}_{(x_{i}, y_{i}) \sim \mathbb{D}^{k}}\mathcal{L}(\theta_{t}; x_{i}, y_{i})$ and $\mathbb{E}_{(x_{i}, y_{i}) \sim \mathbb{D}^{k}}\mathcal{L}(\theta_{t}; \hat{x}_{i}^{*}, y_{i})$ to update the weights $\theta_{t}$, until the total adversarial examples $\hat{x}_{i}^{*}$ are obtained. 

\subsection{Concurrent Adversarial Learning for Large-Batch Training}
\label{section3.3}

As mentioned in the previous section, the vanilla DisAdv requires two sequential gradient computations at each step, where the first gradient computation is to obtain $\hat{x}_i^*$ based on $\mathcal{L}(\theta_t, x_i, y_i)$ and then compute the gradient of $\mathcal{L}(\theta_t, \hat{x}_i^*, y_i)$ to update $\theta_t$. Due to the sequential update nature, this overhead cannot be reduced even when increasing number of processors --- with infinite number of processors the speed of two sequential computations will be twice of one parallel update. This   makes adversarial learning unsuitable for large-batch training. In the following, we propose a simple but novel method to resolve this issue, and provide theoretical analysis on the proposed method. 

% {\color{gray}
% DisAdv is always time  consuming due to the computing of gradients on input data. 
% If we have collected the adversarial data $x_{i}^{*}$ before calculating the loss function $\mathcal{L}_{c}^{k}(\theta_{t}; x_{i}, y_{i})$ and $\mathcal{L}_{a}^{k}(\theta_{t}; x_{i}^{*}, y_{i})$, the problem can be solved. 
% Asynchronous stochastic gradient descent (ASGD) is an critical area in distributed systems, especially in the distributed machine learning, which use delay gradient :

% \begin{equation}
%     \theta_{t+1} = \theta_{t} + \eta g_{t-\tau}(\theta_{t-\tau}),
% \end{equation}

% where 
% % $w_{t}$ denotes the weights of time $t$, 
% $\eta$ represents the learning rate and $g(\theta_{t-\tau})$ is the delay gradient and the delay time is $\tau$. The main idea is to use the delay gradient to update the current weights in large-scale distributed systems. Inspired by the update role of ASGD, we propose a novel asynchronous  training based concurrent adversarial learning to collect adversarial examples and train the model. }

\begin{figure}[t]
\begin{minipage}[c]{0.5\textwidth}
% \begin{minipage}[t]{.5\linewidth}
\centering
\includegraphics[width =\textwidth]{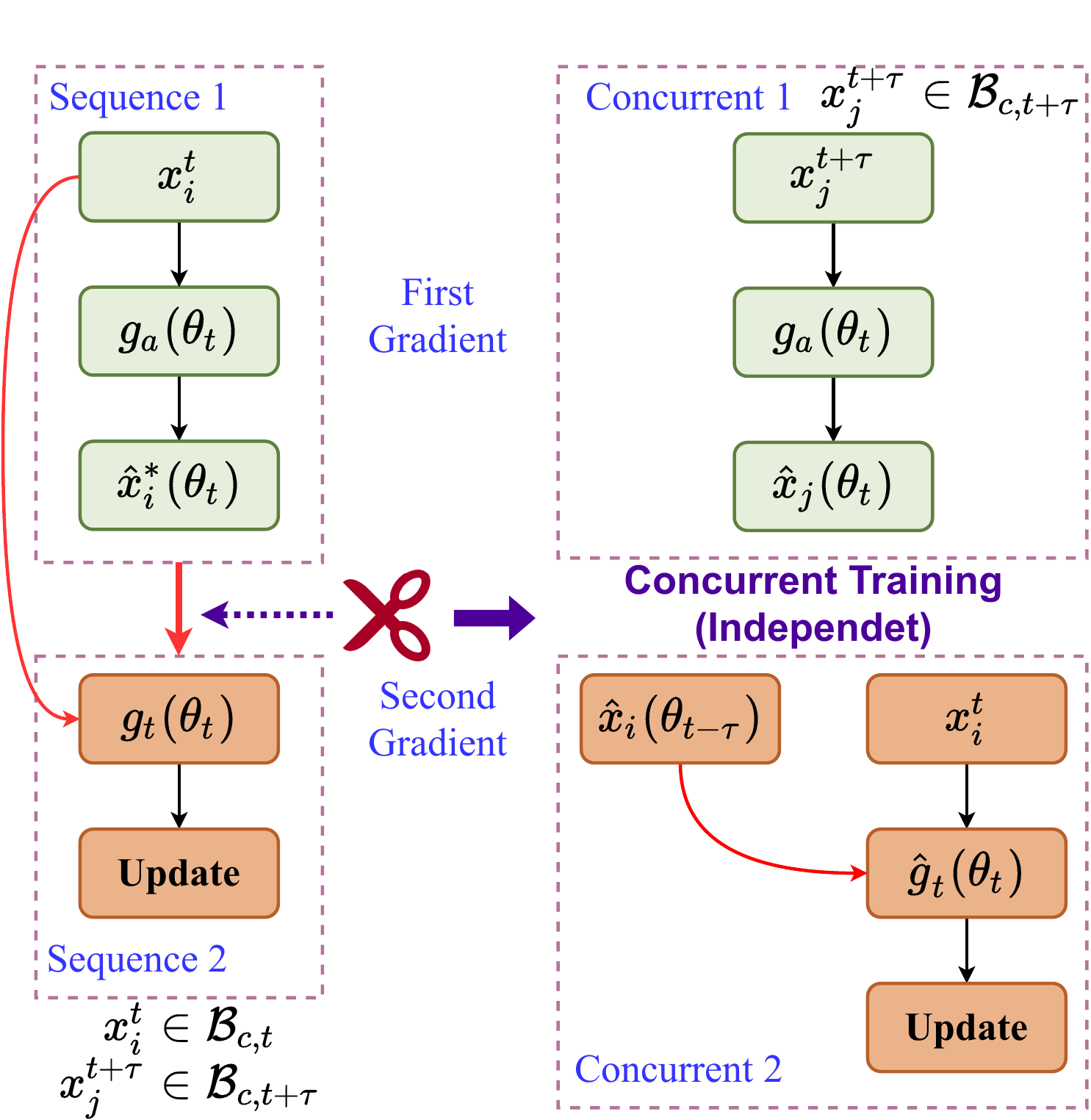}  
\caption{Vanilla Training and Concurrent Training}
\label{fig:main-idea}
% \end{minipage}
\end{minipage}
\hfill
\begin{minipage}[c]{0.47\textwidth}

\begin{algorithm}[H]
  \caption{ConAdv}
  \label{alg:ConcurrentAdv}
\begin{algorithmic}
  \FOR{$t = 1,\cdots, T$}
%   \FOR{each worker $ k \in K$}
  \FOR{$x_{i} \in \mathcal{B}_{c,t}^{k}$}
  \STATE Compute Loss:
  \STATE  $\mathcal{L}(\theta_{t}; x_{i}, y_{i})$ using main BN, \\
  \STATE    $\mathcal{L}_{a}^{k}(\theta_{t};\hat{x}_{i}(\theta_{t-\tau}), y_{i})$ using adv BN, \\
  \STATE 
  $\mathcal{L}_{B}(\theta_{t}) = \mathbb{E}_{\mathcal{B}_{c,t}^{k}}\mathcal{L}(\theta_{t};x_{i},y_{i}) + $\\
   \quad \qquad \quad $\mathbb{E}_{\mathcal{B}_{a,t}^{k}}(\hat{x}_{i}(\theta_{t-\tau}), y_{i})$  \\
  \STATE  Minimize the $\mathcal{L}_{B}(\theta_{t})$ and obtain $g_{t}^{k}(\theta_{t})$ \\
  \ENDFOR
  \FOR{$x_{i}\in\mathcal{B}_{c,t+\tau}^{k}$}
  \STATE  Calculate adv gradient $g_{a}^{k}(\theta_{t})$on $\mathcal{B}_{c,t+\tau}^{k}$ \\
  \STATE  Obtain adv examples $(\hat{x}_{i}(\theta_{t}), y_{i})$ \\
  \ENDFOR 
  \ENDFOR
\STATE Aggregate: $\hat{g}_{t}(\theta_{t}) = \frac{1}{K} \sum_{k=1}^{K}\hat{g}_{t}^{k}(\theta_{t})$ \\
Update weight $\theta_{t+1}$ on parameter sever\\

\end{algorithmic}
\end{algorithm}
\end{minipage}
\end{figure}

\textbf{Concurrent Adversarial Learning (ConAdv) } 
As shown in Figure \ref{fig:main-idea}, our main finding is that  if we use  staled weights ($\theta_{t-\tau}$) for generating adversarial examples, then two sequential computations can be de-coupled and the parameter update step run {\bf concurrently} with the future adversarial example generation step.

Now we formally define the ConAdv procedure. Assume $x_i$ is sampled at iteration $t$, instead of the current weights $\theta_t$, we use staled weights $\theta_{t-\tau}$ (where $\tau$ is the delay) to calculate the gradient and further obtain an approximate adversarial example  $\hat{x}_{i}(\theta_{t-\tau})$:
\begin{equation}
    g_{a}(\theta_{t-\tau}) = \nabla_{x}\mathcal{L}(\theta_{t-\tau}; x_{i}, y_{i}) , \quad
    \hat{x}_{i}(\theta_{t-\tau}) = x_{i} + \alpha \cdot g_{a}(\theta_{t-\tau}).
    %\nabla_{x} \mathcal{L}(\theta_{t-\tau}; x_{i}, y_{i}) 
\end{equation}
%{\color{blue}(cho: check if we need to define $g_a(\theta_{t-\tau})$)}. 
%where $\hat{y}_{i}$ represents the adversarial label since we use  {\color{blue}(cho: what is $\hat{y}$?)}

In this way, 
we can obtain the adversarial sample $\hat{x}_{i}(\theta_{t-\tau})$ through stale weights before updating the model at each step $t$. %{\color{blue}(cho: if we need the $\lambda$ solution onl y in the theoretical analysis, defer this definition to theoretical analysis.)}\textcolor{green}{Done}
Therefore, the training efficiency can be improved. 
%In this paper, we use Projected Gradient Descent (PGD) as the baseline algorithm and propose a novel fast adversarial learning algorithm: Concurrent Adversarial Learning Algorithm (ConAdv). 
The structure of ConAdv as shown in Figure \ref{fig: structure}:
At each step $t$, each worker $k$ can directly concatenate the clean mini-batch data and adversarial mini-batch data to calculate the gradient $\hat{g}_{t}^{k}(\theta_{t})$ and update the model. That is because we have obtain the approximate adversarial example $\hat{x}_{i}$ based on the stale weights $\theta_{t-\tau}$ before iteration $t$. 

In practice, we set $\tau=1$ so the adversarial examples   $\hat{x}_i$ is computed at iteration $t-1$. Therefore, each iteration will compute the current weight update and the adversarial examples for the next batch: 
\begin{equation}
    \theta_{t+1} = \theta_{t} + \frac{\eta}{2} \nabla_{\theta}(\mathbb{E}_{(x_{i},y_{i})\sim\mathcal{B}_{t,c}}\mathcal{L}(\theta_{t};x_{i}, y_{i}) + \mathbb{E}_{\hat{x}_{i}, y_{i}\sim\mathcal{B}_{t,a}}\mathcal{L}(\theta_{t}, \hat{x}_{i}(\theta_{t-1}), y_{i})),
\end{equation}
\begin{equation}
    \hat{x}_{i}(\theta_{t}) = x_{i} + \alpha \cdot \nabla_{x}\mathcal{L}(\theta_{t};x_{i}, y_{i}), \quad \text{ where } \quad (x_{i}, y_{i}) \in \mathcal{B}_{c,t+1}, 
\end{equation}
where $\mathcal{B}_{c,t}=\cup_{k=1}^{k=K}\mathcal{B}_{c,t}^{k}$ denotes clean mini-batch of all workers and $\mathcal{B}_{a,t}=\cup_{k=1}^{k=K}\mathcal{B}_{a,t}^{k}$ represents  adversarial mini-batch of all workers.
These two computations can be parallelized so there is no longer 2 sequential computations at each step. In the large-batch setting 
when the number of workers reach the limit that each batch size can use, 
ConAdv is similarly fast as standard optimizers such as SGD or Adam. 
%when number of processors is much larger than the
 The pseudo code of proposed ConAdv is shown in Algorithm \ref{alg:ConcurrentAdv}.

\subsection{Convergence Analysis}
\label{section 3.4}

In this section, we will show that despite using staled gradient, ConAdv still enjoys nice convergence properties. 
%will prove the  convergence analysis of ConAdv. 
For simplicity, we will use $\mathcal{L}(\theta, x_i)$ as a shorthand for $\mathcal{L}(\theta;x_{i}, y_{i})$ and $\|\cdot\|$ indicates the $\ell_2$ norm. Let optimal adversarial example $x_{i}^{*} = \arg \max_{x_{i}^{*} \in \mathcal{X}_{i}} \mathcal{L}(\theta_{t}, x_{i}^{*})$.
In order to present our main theorem, we will need the following assumptions. 
%{\color{blue}(Cho: I think we also need to define $x^*$ here, is it one gradient step or is it the optimal adv example? (or if the proof can work for both them we can also say that))}\textcolor{green}{Done}

% \textcolor{blue}{change $\|\cdot\|_2$ to $\|\cdot\|$ (and use $ \| \cdot \|$ instead of $||\cdot||$)}

\begin{assumption}
\label{assumption_1}
 The function $\mathcal{L}(\theta,x)$ satisfies the Lipschitzian conditions:
\begin{equation}
\begin{split}
    % \sup_{x} 
    \| \nabla_{x}\mathcal{L}(\theta_{1}; x) - \nabla_{x}\mathcal{L}(\theta_{2}; x) \| \leq L_{x\theta}\|\theta_{1} - \theta_{2}\|, 
% \end{equation}
% \begin{equation}
    % \sup_{x} 
    \| \nabla_{\theta} \mathcal{L}(\theta_{1}; x) - \nabla_{\theta} \mathcal{L}(\theta_{2}; x) \| \leq L_{\theta \theta} \|\theta_{1} - \theta_{2} \|, \\
% \end{equation}
% \begin{equation}
    % \sup_{\theta} 
    \| \nabla_{\theta} \mathcal{L}(\theta; x_{1}) - \nabla_{\theta} \mathcal{L}(\theta; x_{2})\| \leq L_{\theta x} \|x_{1}-x_{2}\|, \
% \end{equation}
% \begin{equation}
    % \sup_{\theta} 
\|\nabla_{x} \mathcal{L}(\theta; x_{1}) - \nabla_{x} \mathcal{L}(\theta; x_{2}) \| \leq L_{xx}\|x_{1} - x_{2}\|.
\end{split}
\end{equation}

\end{assumption}

\begin{assumption}
\label{assumption_2}
 $\mathcal{L}(\theta,x)$ is locally  $\mu$-strongly concave in $\mathcal{X}_{i} = \{x^{*}: ||x^{*} - x_{i}||_{\infty} \leq \epsilon \}$ for all $i \in [n]$, i.e., for any $x_{1}, x_{2} \in \mathcal{X}_{i}$,
\begin{equation}
    \mathcal{L}(\theta, x_{1}) \leq \mathcal{L}(\theta, x_{2}) + \langle \nabla_{x}\mathcal{L}(\theta, x_{2}), x_{1} - x_{2} \rangle - \frac{\mu}{2} \| x_{1} - x_{2} \|,
\end{equation}
\end{assumption}
Assumption 2 can be verified based on the relation between robust optimization and distributional robust optimization in \cite{sinha2017certifiable,lee2017minimax}.

%\textcolor{blue}{(cho: the local strongly convex condition could be unrealistics? If previous paper also used this we can explicitly point out.)}\textcolor{green}{Done}

\begin{assumption}
 The concurrent stochastic gradient $\hat{g}(\theta_{t}) = \frac{1}{2|\mathcal{B}|}\sum_{i=1}^{|\mathcal{B}|}(\nabla_{\theta}\mathcal{L}(\theta_{t};x_{i}) + \nabla_{\theta}\mathcal{L}(\theta_{t}, \hat{x}_{i}))$ is bounded by the constant $M$:
\begin{equation}
    \|\hat{g}(\theta_{t})\| \leq M.
\end{equation}
\end{assumption}

\begin{assumption}
  Suppose $\mathcal{L}_{D}(\theta_{t}) = \frac{1}{2n}\sum_{i=1}^{n}(\mathcal{L}(\theta_{t}, x_{i}^{*}) + \mathcal{L}(\theta_{t}, x_{i}))$, $g(\theta_{t}) = \frac{1}{2|\mathcal{B}|}\sum_{i=1}^{|\mathcal{B}|}(\nabla_{\theta}\mathcal{L}(x_{i}) + \nabla_{\theta}\mathcal{L}(\theta_{t}, x_{i}^{*}))$ and $\mathbb{E}[g(\theta_{t})] = \nabla \mathcal{L}_{D}(\theta_{t})$, where $|\mathcal{B}|$ represents batch size . The variance of $g(\theta_{t})$ is bounded by $\sigma^{2}$:
\begin{equation}
    \mathbb{E}[\|g(\theta_{t}) - \nabla \mathcal{L}_{D}(\theta_{t})\|^{2}] \leq \sigma^{2}.
\end{equation}

\end{assumption}

Based on the above assumptions, we can obtain the upper bound between original  adversarial example $x_{i}^{*}(\theta_{t})$ and  concurrent adversarial example $x_{i}^{*}(\theta_{t-\tau})$, where $\tau$ is the delay time.

\begin{lemma}
Under Assumptions 1 and 2, we have 
\begin{equation}
\|x_{i}^{*}(\theta_{t}) - x_{i}^{*}(\theta_{t-\tau})|| \leq \frac{L}{\mu}||\theta_{t} - \theta_{t-\tau}\| \leq \frac{L_{x \theta}}{\mu} \eta \tau M.
\end{equation}

% where $x_{i}^{*}(\theta_{t})$ and $x_{i}^{*}(\theta_{t-\tau})$ denote the adversarial example of $x_{i}$ calculated by $\theta_{t}$ and $\theta_{t-\tau}$, respectively.

% \textbf{Proof}: Please see Appendix 1.

\end{lemma}

Lemma 1 illustrates the relation between $x_{i}^{*}(\theta_{t})$ and $x_{i}^{*}(\theta_{t-\tau})$, which is bounded by the delay $\tau$. When the delay is small enough,  $x_{i}^{*}(\theta_{t-\tau})$ can be regarded as an approximator of $x_{i}^{*}(\theta_{t})$. We now establish the convergence rate as the following theorem.

\begin{theorem}

Suppose Assumptions 1, 2, 3 and 4 hold. Let loss function $ \mathcal{L}_{D}(\theta_{t}) = \frac{1}{2n}\sum_{i=1}^{n}(\mathcal{L}(\theta_{t}; x_{i}^{*}, y_{i}) + \mathcal{L}(\theta_{t}; x_{i}, y_{i}))$ and  $\hat{x}_{i}(\theta_{t-\tau})$ be the $\lambda$-solution of $x_{i}^{*}(\theta_{t- \tau})$: $  \langle   x_{i}^{*}(\theta_{t-\tau}) - \hat{x}_{i}(\theta_{t-\tau}), \nabla_{x}\mathcal{L}(\theta_{t-\tau}; \hat{x}_{i}(\theta_{t-\tau}), y_{i}) \rangle \leq \lambda $. Under Assumptions 1 and 2, for the concurrent stochastic gradient $\hat{g}(\theta)$.
%\textcolor{blue}{(Cho: I think $L_S$ should be robust plus clean loss? Here it seems you only have clean loss? And what does $S$ mean?)} \textcolor{green}{Done}
If the step size of outer minimization is set to $\eta_{t} = \eta = \min(1/L, \sqrt{\Delta / L \sigma^{2}T})$. Then the output of Algorithm 1 satisfies:

\begin{equation}
        \frac{1}{T}\sum_{t=0}^{T-1}\mathbb{E}[||\nabla \mathcal{L}_{D}(\theta_{t})||_{2}^{2}] \leq 2\sigma \sqrt{\frac{L\Delta}{T}} + \frac{L_{\theta x}^{2} }{2}(\frac{\tau M L_{x \theta}}{L \mu} + \sqrt{\frac{\lambda}{\mu}})^{2}
\end{equation}

where $L=L_{\theta \theta} +  \frac{L_{x \theta}}{2\mu}L_{\theta x}$

% \textbf{Proof}: Please see Appendix 4.

\end{theorem}

Our result provides a formal convergence rate of  ConAdv and it can converge to a first-order stationary point at a sublinear rate up to a precision of $\frac{L_{\theta x}^{2} }{2}(\frac{\tau M L_{x \theta}}{L \mu} + \sqrt{\frac{\lambda}{\mu}})^{2}$, which is related to $\tau$.  In practice we use the smallest delay $\tau=1$ as discussed in the previous subsection. 
%In addition, we can find that the convergence rate is related to the delay $\tau$. When $\tau$ is small enough, the delay weight $\theta_{t-\tau}$ doesn't have   a significant effect for the convergence. Therefore, we set $\tau=1$ in all experiments since that is enough for collecting adversarial examples before updating the model.

\section{Experimental Results}
\label{section4.1}

% In this section, we conduct large-batch training experiments for ResNet on ImageNet and provide several further analysis on ConvAdv and data augmentation. 

\subsection{Experimental Setup}
% \subsubsection{Ar}
\textbf{Architectures and Datasets.}
We select ResNet as our default architectures. More specially, we use the mid-weight version (ResNet-50) to evaluate the performance of our proposed algorithm. The dataset we used  in this paper is ImageNet-1k, which consists of 1.28 million images for training and 50k images for testing. The convergence baseline of ResNet-50 in MLPerf is 75.9\% top-1 accuracy in 90 epochs (i.e. ResNet-50 version 1.5 \cite{goyal2017accurate}). 

% All experiments are conducted 5 times from different starting points to ensure the reliable results.

% \textbf{Scaling up Batch Size}

\textbf{Implementation Details.}
We use TPU-v3 for all our experiments and the same setting as the baseline. %For ResNet-50 on ImageNet-1k,
We consider 90-epoch training for ResNet-50. 
%set the number of epochs as 90 and the number of epochs is 350 for EfficientNet. 
For data augmentation, we mainly consider AutoAug (AA). In addition, we use LARS \cite{you2017scaling} to train all the models. Finally, for adversarial training, we always use 1-step PGD attack with random initialization.

% The critical method we use:

\begin{figure}[htbp]
\centering                                      %居中
\subfigure[ResNet-50]{          %第一张子图
\begin{minipage}[t]{0.36\linewidth}
\centering                                  %子图居中
\includegraphics[width=\linewidth]{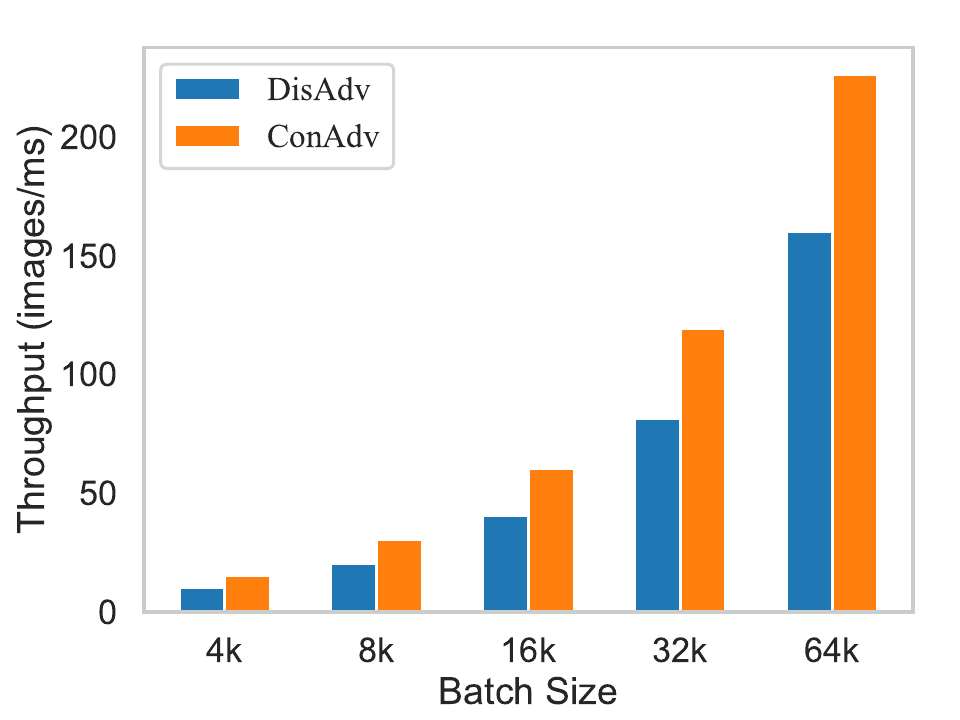} 
\end{minipage}}\quad \quad \quad
\subfigure[ResNet-50 Limit]{          
\begin{minipage}[t]{0.36\linewidth}
\centering                              
\includegraphics[width=\linewidth]{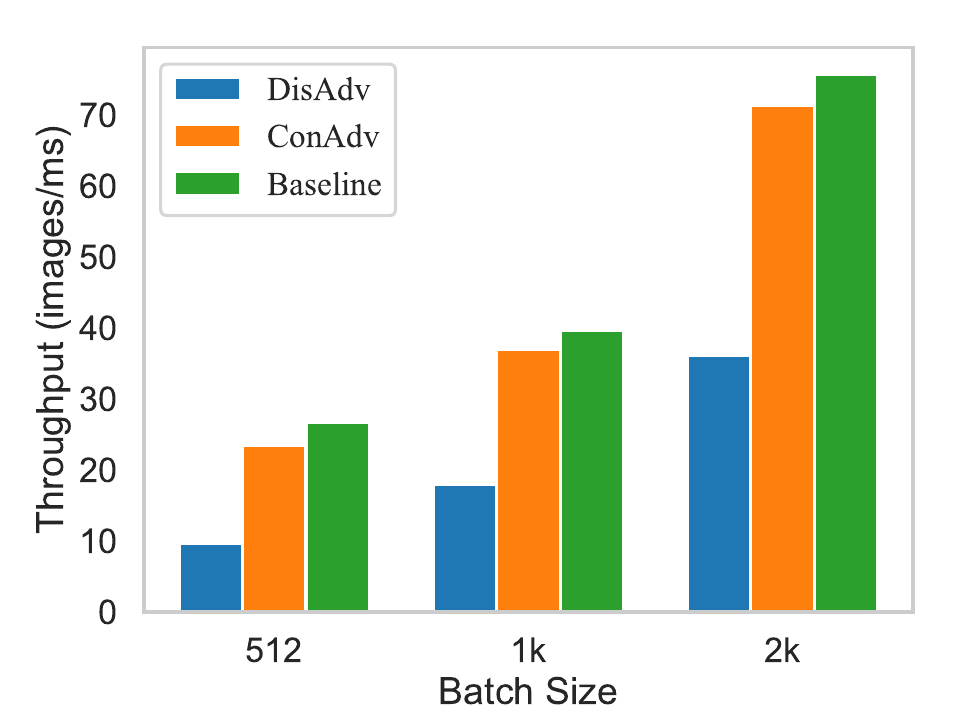}       
\end{minipage}}
\caption{ (a): throughput on scaling up batch size for ResNet-50, (b): throughtput when the number of processors reach the limit that each batch size can use for ResNet-50 .} %              
\label{fig: analysis_speed} 
\end{figure}

\subsection{ImageNet Training with ResNet}
\label{section4.2}

% \begin{wrapfigure} {1}{0.4\textwidth}
% \includegraphics[width = 0.4\textwidth]{vi_speed.pdf}  
% \caption{Throughput-Batch Size}
% \label{fig:vi_speed}
% \end{wrapfigure}

We train ResNet-50 with ConAdv and compare it with vanilla training and DisAdv.
% Firstly, Let us use the example ImageNet training with ResNet-50 to illustrate our proposed algorithm. 
%We try to evaluate the performance of proposed concurrent adversarial learning algorithm based on ResNet-50 model. 
The experimental results of scaling up batch size in Table \ref{resnet-table-accuracy} illustrates that ConAdv can obtain the similar accuracy compared with DisAdv and meanwhile speeding up the training process. More specially, we can find that the top-1 accuracy of all methods are stable when the batch size is increased from 4k to 32k. After that, the performance starts to drop, which illustrates the bottleneck of large-batch training.
%because the number of iterations is reduced. 
However, ConAdv can improve the top-1 accuracy and the improved performance is stable as DisAdv does when the batch size reaches the bottleneck (such as 32k, 64k, 96k), but AutoAug gradually reaches its limitations. For instance, the top-1 accuracy increases from 74.3 to 75.3 when using ConAdv with a batch size of 96k and improved accuracy is $0.7\%$, $0.9\%$ and $1.0\%$ for 32k, 64k and 96k. However, AutoAug cannot further improve the top-1 accuracy when batch size is 96k. The above results illustrate that adversarial learning can successfully maintain a good test performance in the large-batch training setting and can outperform data augmentation. \\

\begin{table}[H]
\setlength{\abovecaptionskip}{-0.2cm}
\setlength{\belowcaptionskip}{-0.2cm}
\renewcommand\arraystretch{1.2}
\small
\caption{Top-1 accuracy for ResNet-50 on ImageNet}
\label{resnet-table-accuracy}
\vskip 0.15in
\begin{center}
\begin{small}
\begin{sc}
\begin{tabular}{lccccccr}

\multicolumn{1}{l}{\bf Method} & \multicolumn{1}{l}{\bf 1k} & \multicolumn{1}{l}{\bf 4k} & \multicolumn{1}{l}{\bf 8k} & \multicolumn{1}{l}{\bf 16k} & \multicolumn{1}{l}{\bf 32k} & \multicolumn{1}{l}{\bf 64k} & \multicolumn{1}{l}{\bf 96k} \\
\hline
\multicolumn{1}{l}{\bf ResNet-50}    & 76.9   & 76.9 & 76.6 & 76.6 & 76.6 & 75.3& 74.3  \\
\multicolumn{1}{l}{\bf ResNet-50+AA} & \bf{77.5} & \bf{77.5} & \bf{77.4} & 77.1 & 76.9 & 75.6& 74.3 \\
\multicolumn{1}{l}{\bf ResNet-50+DisAdv} & 77.4  & 77.4  & \bf{77.4} & \bf{77.4}      & \bf{77.3} & \bf{76.2} & \bf{75.3} \\
\multicolumn{1}{l}{\bf ResNet-50+ConAdv} &  77.4  & 77.4 & \bf{77.4}   & \bf{77.4} &  \bf{77.3} & \bf{76.2} & \bf{75.3}\\
\hline
\end{tabular}
\end{sc}
\end{small}
\end{center}
\vskip -0.1in
\end{table}

%large-batch cannot hurt the performance of adversarial learning and adversarial learning can further expand batch-size of large-batch training. 

% \begin{wrapfigure} {1}{0.4\textwidth}
% \includegraphics[width = 0.4\textwidth]{vi-limit-resnet.pdf}  
% \caption{Throughput-Batch Size}
% \label{fig:vi_speed_limit_resnet}
% \end{wrapfigure}

In addition, Figure \ref{fig: analysis_speed}(a) presents the throughput (images/ms) on scaling up batch size. We can observe that ConAdv can further increase throughput and  accelerate the training process. 
To obtain accurate statistics of BatchNorm, we need to make sure each worker has at least 64 examples to calculate them (Normal Setting). Thus, the number of cores is [Batch Size / 64].
For example, we use TPU v3-256  to train DisAdv when batch size is 32k, which has 512 cores (32k/64=512). As shown in Figure \ref{fig: analysis_speed}(a), the throughput of DisAdv increases from 10.3 on 4k to 81.7 on 32k and CondAdv achieve about 1.5x speedup compared with DisAdv, which verifies our proposed ConAdv can maintain the accuracy of large-batch training and meanwhile accelerate the training process.

% In addition, Distributed BatchNorm can aggregate the statistics of multiple workers and we can use more machines to speedup the training process. 
To simulate the speedup when the number of workers reach the limit that each Batch Size can use, we use a large enough distributed system to train the model with the batch size of 512, 1k and 2k on TPU v3-128, TPU v3-256 and TPU v3-512 , respectively. The result is shown in Figure \ref{fig: analysis_speed}(b), we can obtain that ConAdv can achieve  about 2x speedup compared with DisAdv. Furthermore, in this scenario we can observe ConAdv can achieve the similar throughput as Baseline (vanilla ResNet-50 training). For example, compared with DisAdv, the throughput increases from 36.1 to 71.3 when using ConAdv with a batch size of 2k. In addition, the throughput is 75.7, which illustrates that ConAdv can achieve a similar speed as baseline. However, ConAdv can expand to larger batch size than baseline. Therefore, ConAdv can further accelerate the training of deep neural network.

\subsection{ImageNet Training with Data Augmentation}

To explore the limit of our method and evaluate whether adversarial learning can be combined with data augmentation for large-batch training, we further apply data augmentation into proposed adversarial learning algorithm and the results are shown in Table \ref{table-resnet-aa}. 
We can find that ConAdv can further improve the performance of large-batch training on ImageNet when combined with Autoaug (AA). 
Under this setting, we can expand the batch size to more than 96k, which can improve the algorithm efficiency and meanwhile benefit for the machine utilization. For instance, for ResNet, the top-1 accuracy increases from 74.3 to 76.2 under 96k when using ConAdv and AutoAug. 
\\

\begin{table}[H]
\setlength{\abovecaptionskip}{-0.2cm}
\setlength{\belowcaptionskip}{-0.2cm}
\renewcommand\arraystretch{1.2}
\caption{Top-1 accuracy with AutoAug on ImageNet}
\label{table-resnet-aa}

\vskip 0.15in
\begin{center}
\begin{small}
\begin{sc}
\begin{tabular}{lccccccc}

\multicolumn{1}{l}{\bf Method} & \multicolumn{1}{c}{\bf 1k} & \multicolumn{1}{c}{\bf 4k} & \multicolumn{1}{c}{\bf 8k} & \multicolumn{1}{c}{\bf 16k} & \multicolumn{1}{c}{\bf 32k} & \multicolumn{1}{c}{\bf 64k} & \multicolumn{1}{c}{\bf 96k} \\
\hline
\multicolumn{1}{l}{\bf ResNet-50}    & 76.9  & 76.9 & 76.6 & 76.6 & 76.6 & 75.3& 74.3  \\
\multicolumn{1}{l}{\bf ResNet-50+AA} & 77.5 & 77.5 & 77.4 & 77.1 & 76.9 & 75.6& 74.3 \\
\multicolumn{1}{c}{\bf ResNet-50+ConAdv+AA} & \bf{78.5} & \bf{78.5} & \bf{78.5} & \bf{78.5} & \bf{78.3} & \bf{77.3} & \bf{76.2} \\
\hline

% \midrule
% EfficientNet-B2    & 79.5  & 79.4 & 79.5 & 79.4 & 79.3 & 78.9& / \\
% EfficientNet-B2+AA & 80.0 & 80.0 & 79.8 & 79.5 & 79.4  & 79.1& / \\
% EfficientNet-B2+ConAdv+AA & \bf{80.4}  & \bf{80.4} & \bf{80.4} & \bf{80.4} & \bf{80.2} & \bf{80.0}& / \\
\hline
\end{tabular}
\end{sc}
\end{small}
\end{center}
\vskip -0.1in
\end{table}

\subsection{Training Time}

The wall clock training times for ResNet-50 are shown in Table \ref{exp_tab_time}. We can find that the training time gradually decreases with batch size increasing. For example, the training time of ConAdv decreases 
% That verifies that large-batch training can improve the training speed and accelerate the training process. 

\begin{wraptable}{r}{6.8cm}
\centering
\caption{Training Time Analysis}
\label{exp_tab_time}
\renewcommand\arraystretch{1.2}
\begin{tabular}{lccc}

\multicolumn{1}{c}{\bf Method} & \multicolumn{1}{c}{\bf 16k} & \multicolumn{1}{c}{\bf 32k} & \multicolumn{1}{c}{\bf 64k}  
\\ \hline
\multicolumn{1}{l}{\bf ResNet-50}    & \bf 1194s   & \bf 622s & /   \\ 
\multicolumn{1}{l}{\bf DisAdv} & 1657s & 1191s & 592s  \\
\multicolumn{1}{l}{\bf ConAdv} & 1277s  & 677s  & \bf 227s \\
\hline
\end{tabular}

\end{wraptable}

% \begin{table}[H]
% \caption{Training Times for ResNet-50 on ImageNet-1k}
% \label{exp_tab_time}
% \vskip 0.15in
% \begin{center}
% \begin{small}
% \begin{sc}
% \begin{tabular}{lccc}

% \multicolumn{1}{c}{\bf Method} & \multicolumn{1}{c}{\bf 16k} & \multicolumn{1}{c}{\bf 32k} & \multicolumn{1}{c}{\bf 64k}  
% \\ \hline
% \multicolumn{1}{l}{\bf Vanilla ResNet-50}    & \bf 1194s   & \bf 622s & /   \\ 
% \multicolumn{1}{l}{\bf DisAdv} & 1657s & 1191s & 592s  \\
% \multicolumn{1}{l}{\bf ConAdv} & 1277s  & 677s  & \bf 227s \\
% \hline
% \end{tabular}
% \end{sc}
% \end{small}
% \end{center}
% \vskip -0.1in
% \end{table}

 from  1277s to 227s when scale the batch size from 16k to 64k. In addition, we can find that DisAdv need about 1.5x training time compared with vanilla ResNet-50 
but ConAdv can efficiently reduce the training of DisAdv to a level similar to vanilla ResNet. For instance, the training time of DisAdv is reduced from 1191s to 677s when using ConAdv. Noted that we don’t report the clock time for vanilla ResNet-50 at 64k since the top-1 accuracy is below the MLPerf standard 75.9\%. The number of machines required to measure the maximum speed at 96K exceeds our current resources. The comparison on 32k and 64k also can evaluate the runtime improvement.

\subsection{Generalization Gap}

To the best of our knowledge, theoretical analysis of generalization errors in large-batch setting is still an open problem. However, we empirically found that our method can successfully reduce the generalization gap in large-batch training. The experimental results in Table \ref{table-resnet-gap} indicate that ConAdv can narrow the generalization gap. For example, the generalization gap is 4.6 for vanilla ResNet-50 at 96k and ConAdv narrows the gap to 2.7. In addition, combining ConAdv with AutoAug, the training accuracy and test accuracy can further increase and meanwhile maintain the similar generalization gap. \\

\begin{table}[H]
\setlength{\abovecaptionskip}{-0.2cm}
\setlength{\belowcaptionskip}{-0.2cm}
\renewcommand\arraystretch{1.2}
% \tiny
\caption{Generalization Gap of Large-Batch Training on ImageNet-1k}
\label{table-resnet-gap}
\vskip 0.15in
\begin{center}
\begin{scriptsize}
% \begin{sc}
\begin{tabular}{l|cccc|cccc|cccc}
\hline
& \multicolumn{4}{c|}{\bf Vanilla ResNet-50} & \multicolumn{4}{c|}{\bf ConAdv} &
\multicolumn{4}{c}{\bf ConAdv + AA} \\
\multicolumn{1}{c|}{} & \multicolumn{1}{c}{\bf 16k} & \multicolumn{1}{c}{\bf 32k} & \multicolumn{1}{c}{\bf 64k} & \multicolumn{1}{c|}{\bf 96k} & \multicolumn{1}{c}{\bf16k} & \multicolumn{1}{c}{\bf 32k} & \multicolumn{1}{c}{\bf 64k} & \multicolumn{1}{c|}{\bf 96k} & \multicolumn{1}{c}{\bf16k} & \multicolumn{1}{c}{\bf 32k} & \multicolumn{1}{c}{\bf 64k} & \multicolumn{1}{c}{\bf 96k} 
\\ \hline 
\multicolumn{1}{l|}{\bf Training Accuracy}      & 81.4   & 82.5    & 79.6    & 78.9 & 80.3   & 80.8   & 78.2   & 78.0 & 81.6   & 81.7   & 79.6   & 78.4   \\
\multicolumn{1}{l|}{\bf Test Accuracy}  & 76.6   & 76.6    & 75.3    & 74.3 & 77.4   & 77.3   & 76.2   & 75.3 & 78.5   & 78.3   & 77.3   & 76.2\\
\multicolumn{1}{l|}{\bf Generalization Gap}  & 4.8    & 5.9     & 4.3     & 4.6 & 2.9   & 3.5   & 2.0   & 2.7 & 3.1   & 3.4   & 2.3   & 2.2 \\
\hline

% \midrule
% EfficientNet-B2    & 79.5  & 79.4 & 79.5 & 79.4 & 79.3 & 78.9& / \\
% EfficientNet-B2+AA & 80.0 & 80.0 & 79.8 & 79.5 & 79.4  & 79.1& / \\
% EfficientNet-B2+ConAdv+AA & \bf{80.4}  & \bf{80.4} & \bf{80.4} & \bf{80.4} & \bf{80.2} & \bf{80.0}& / \\
\hline
\end{tabular}
% \end{sc}
\end{scriptsize}
\end{center}
\vskip -0.1in
\end{table}

\subsection{Analysis of Adversarial perturbation}

Adversarial learning calculate an adversarial perturbation on input data to smooth the decision boundary and help the model converge to the flat minima. In this section, we analyze the effects of different perturbation values for the performance of large-batch training on ImageNet. The analysis results are illustrated in Table \ref{Exp_adv_p1}. It presents that we should increase the attack intensity as the batch size increasing. For example, the best attack perturbation value increases from 3 (32k) to 7 (96k) for ResNet-50 and from 8 (16k) to 12 (64k). In addition, we should increase the perturbation value when using data augmentation. For example, the perturbation value should be 3 for original ResNet-50 but be 5 when data augmentation is applied. \\

% \begin{table}[t]
% \caption{Sample table title}
% \label{sample-table}
% \begin{center}
% \begin{tabular}{ll}
% \multicolumn{1}{c}{\bf PART}  &\multicolumn{1}{c}{\bf DESCRIPTION}
% \\ \hline \\
% Dendrite         &Input terminal \\
% Axon             &Output terminal \\
% Soma             &Cell body (contains cell nucleus) \\
% \end{tabular}
% \end{center}
% \end{table}

\begin{table}[H]
\setlength{\abovecaptionskip}{-0.2cm}
\setlength{\belowcaptionskip}{-0.5cm}
\renewcommand\arraystretch{1.2}

\caption{Experiment Results (Top-1 Accuracy) when useing Different Adversarial Perturbation.}
\label{Exp_adv_p1}
\vskip 0.15in
\begin{center}
\begin{small}
\begin{sc}
\resizebox{\textwidth}{!}{
\begin{tabular}{lccccccccccccc}
\multicolumn{1}{c}{\bf Method} & \multicolumn{1}{c}{\bf Batch Size} & \multicolumn{1}{c}{\bf p=0} & \multicolumn{1}{c}{\bf p=1} & \multicolumn{1}{c}{\bf p=2} & \multicolumn{1}{c}{\bf p=3} & \multicolumn{1}{c}{\bf p=4} & \multicolumn{1}{c}{\bf p=5} & \multicolumn{1}{c}{\bf p=6} & \multicolumn{1}{c}{\bf p=7} & \multicolumn{1}{c}{\bf p=8} & \multicolumn{1}{c}{\bf p=9} & \multicolumn{1}{c}{\bf p=10} & \multicolumn{1}{c}{\bf p=12}\\
\hline 
\multicolumn{1}{l}{\bf ResNet-50 + ConAdv}& 32k& 76.8 & 77.2& 77.3& \bf{77.4} & 77.3& 77.3& 77.3& 77.3& 77.3& 77.3& 77.2& 77.2\\
\multicolumn{1}{l}{\bf ResNet-50 + ConAdv + AA}& 32k& 77.8& 78.0& 78.1& 78.1& 78.0& \bf{78.3}& 78.2& 78.2& 78.2& 78.2& 78.2& 78.1\\
\multicolumn{1}{l}{\bf ResNet-50 + ConAdv}& 64k&  75.7& 76.2& 76.3& 76.3& 76.4& \bf{76.7}& 76.4& 76.4& 76.4& 76.4& 76.4& 76.3\\
\multicolumn{1}{l}{\bf ResNet-50 + ConAdv + AA}& 64k& 76.8& 77.0& 76.8& 77.0& 77.1& 77.1& 77.2& \bf{77.4} & 77.2& 77.1& 77.1& 77.1\\
\multicolumn{1}{l}{\bf ResNet-50 + ConAdv} & 96k&  74.6& 75.1& 75.1& 75.1& 75.3& 75.1& 75.1& 75.1& \bf{75.3}& 75.2&     75.1& 75.1\\
\multicolumn{1}{l}{\bf ResNet-50 + ConAdv + AA}& 96k& 75.8 & 75.9& 75.8& 76.0& 76.0& 76.0& 76.0& 76.1& \bf{76.2} & \bf{76.2}&  76.0 & 76.0\\
\hline
\end{tabular}}
\end{sc}
\end{small}
\end{center}
\end{table}
\section{Conclusions}
\label{section5.1}

We firstly analyze the effect of data augmentation for large-batch training and propose a novel distributed adversarial learning algorithm to scale to larger batch size. To reduce the overhead of adversarial learning, we further propose a novel concurrent adversarial learning to decouple the two sequential gradient computations in adversarial learning. 
%improve the efficiency of adversarial learning. 
We evaluate our proposed method on ResNet. 
The experimental results show that our proposed method is beneficial for large-batch training.
\section{Acknowledgements}

We thank Google TFRC for supporting us to get access to the Cloud TPUs. We thank CSCS (Swiss National Supercomputing Centre) for supporting us to get access to the Piz Daint supercomputer. We thank TACC (Texas Advanced Computing Center) for supporting us to get access to the Longhorn supercomputer and the Frontera supercomputer. We thank LuxProvide (Luxembourg national supercomputer HPC organization) for supporting us to get access to the MeluXina supercomputer.

% \section{Ethics Statement}

% We do not have any potential ethics issues in this paper. We hope to propose a novel distributed adversarial learning algorithm to accelerate large-batch training.

% \section{Reproducibility Statement}

% We provide our code in Appendix \ref{Appendix:5}. In addition, we also list our main hyperparameters for large-batch training in Appendix \ref{Appendix:5} (Table \ref{resnet_hyperparam}). For experimental details, we introduce our experiment settings in section \ref{section4.1}, such as the dataset, model architecture, data augmentation, optimizer and so on.

\bibliography{reference}

\begin{thebibliography}{43}
\providecommand{\natexlab}[1]{#1}
\providecommand{\url}[1]{\texttt{#1}}
\expandafter\ifx\csname urlstyle\endcsname\relax
  \providecommand{\doi}[1]{doi: #1}\else
  \providecommand{\doi}{doi: \begingroup \urlstyle{rm}\Url}\fi

\bibitem[Akiba et~al.(2017)Akiba, Suzuki, and Fukuda]{akiba2017extremely}
Takuya Akiba, Shuji Suzuki, and Keisuke Fukuda.
\newblock Extremely large minibatch sgd: Training resnet-50 on imagenet in 15
  minutes.
\newblock \emph{arXiv preprint arXiv:1711.04325}, 2017.

\bibitem[Andriushchenko \& Flammarion(2020)Andriushchenko and
  Flammarion]{andriushchenko2020understanding}
Maksym Andriushchenko and Nicolas Flammarion.
\newblock Understanding and improving fast adversarial training.
\newblock In \emph{Advances in Neural Information Processing Systems}, 2020.

\bibitem[Berman et~al.(2019)Berman, J{\'e}gou, Vedaldi, Kokkinos, and
  Douze]{berman2019multigrain}
Maxim Berman, Herv{\'e} J{\'e}gou, Andrea Vedaldi, Iasonas Kokkinos, and
  Matthijs Douze.
\newblock Multigrain: a unified image embedding for classes and instances.
\newblock \emph{arXiv preprint arXiv:1902.05509}, 2019.

\bibitem[Chen et~al.(2021)Chen, Xie, Tan, Zhang, Hsieh, and
  Gong]{chen2021robust}
Xiangning Chen, Cihang Xie, Mingxing Tan, Li~Zhang, Cho-Jui Hsieh, and Boqing
  Gong.
\newblock Robust and accurate object detection via adversarial learning, 2021.

\bibitem[Cheng et~al.(2021)Cheng, Gan, Cheng, Wang, Hsieh, and
  Liu]{cheng2021adversarial}
Minhao Cheng, Zhe Gan, Yu~Cheng, Shuohang Wang, Cho-Jui Hsieh, and Jingjing
  Liu.
\newblock Adversarial masking: Towards understanding robustness trade-off for
  generalization, 2021.
\newblock URL \url{https://openreview.net/forum?id=LNtTXJ9XXr}.

\bibitem[Codreanu et~al.(2017)Codreanu, Podareanu, and
  Saletore]{codreanu2017scale}
Valeriu Codreanu, Damian Podareanu, and Vikram Saletore.
\newblock Scale out for large minibatch sgd: Residual network training on
  imagenet-1k with improved accuracy and reduced time to train.
\newblock \emph{arXiv preprint arXiv:1711.04291}, 2017.

\bibitem[Cubuk et~al.(2019)Cubuk, Zoph, Mane, Vasudevan, and
  Le]{Cubuk_2019_CVPR}
Ekin~D. Cubuk, Barret Zoph, Dandelion Mane, Vijay Vasudevan, and Quoc~V. Le.
\newblock Autoaugment: Learning augmentation strategies from data.
\newblock In \emph{IEEE Conference on Computer Vision and Pattern Recognition},
  2019.

\bibitem[Deng et~al.(2009)Deng, Dong, Socher, Li, Li, and
  Fei-Fei]{imagenet_cvpr09}
J.~Deng, W.~Dong, R.~Socher, L.-J. Li, K.~Li, and L.~Fei-Fei.
\newblock {ImageNet: A Large-Scale Hierarchical Image Database}.
\newblock In \emph{IEEE Conference on Computer Vision and Pattern Recognition},
  2009.

\bibitem[Devarakonda et~al.(2017)Devarakonda, Naumov, and
  Garland]{devarakonda2017adabatch}
Aditya Devarakonda, Maxim Naumov, and Michael Garland.
\newblock Adabatch: Adaptive batch sizes for training deep neural networks.
\newblock \emph{arXiv preprint arXiv:1712.02029}, 2017.

\bibitem[Devlin et~al.(2019)Devlin, Chang, Lee, and Toutanova]{DevlinCLT19}
Jacob Devlin, Ming{-}Wei Chang, Kenton Lee, and Kristina Toutanova.
\newblock {BERT:} pre-training of deep bidirectional transformers for language
  understanding.
\newblock In Jill Burstein, Christy Doran, and Thamar Solorio (eds.),
  \emph{Conference of the North American Chapter of the Association for
  Computational Linguistics: Human Language Technologies, {NAACL-HLT}}, 2019.

\bibitem[Goodfellow et~al.(2015)Goodfellow, Shlens, and
  Szegedy]{goodfellow2014explaining}
Ian Goodfellow, Jonathon Shlens, and Christian Szegedy.
\newblock Explaining and harnessing adversarial examples.
\newblock In \emph{International Conference on Learning Representations}, 2015.

\bibitem[Goyal et~al.(2017)Goyal, Doll{\'a}r, Girshick, Noordhuis, Wesolowski,
  Kyrola, Tulloch, Jia, and He]{goyal2017accurate}
Priya Goyal, Piotr Doll{\'a}r, Ross Girshick, Pieter Noordhuis, Lukasz
  Wesolowski, Aapo Kyrola, Andrew Tulloch, Yangqing Jia, and Kaiming He.
\newblock Accurate, large minibatch sgd: Training imagenet in 1 hour.
\newblock \emph{arXiv preprint arXiv:1706.02677}, 2017.

\bibitem[He et~al.(2016)He, Zhang, Ren, and Sun]{He_2016_CVPR}
Kaiming He, Xiangyu Zhang, Shaoqing Ren, and Jian Sun.
\newblock Deep residual learning for image recognition.
\newblock In \emph{IEEE Conference on Computer Vision and Pattern Recognition},
  2016.

\bibitem[Hoffer et~al.(2017)Hoffer, Hubara, and Soudry]{hoffer2017train}
Elad Hoffer, Itay Hubara, and Daniel Soudry.
\newblock Train longer, generalize better: closing the generalization gap in
  large batch training of neural networks.
\newblock \emph{arXiv preprint arXiv:1705.08741}, 2017.

\bibitem[Iandola et~al.(2016)Iandola, Moskewicz, Ashraf, and
  Keutzer]{iandola2016firecaffe}
Forrest~N Iandola, Matthew~W Moskewicz, Khalid Ashraf, and Kurt Keutzer.
\newblock Firecaffe: near-linear acceleration of deep neural network training
  on compute clusters.
\newblock In \emph{IEEE Conference on Computer Vision and Pattern Recognition},
  2016.

\bibitem[Jia et~al.(2018)Jia, Song, He, Wang, Rong, Zhou, Xie, Guo, Yang, Yu,
  et~al.]{jia2018highly}
Xianyan Jia, Shutao Song, Wei He, Yangzihao Wang, Haidong Rong, Feihu Zhou,
  Liqiang Xie, Zhenyu Guo, Yuanzhou Yang, Liwei Yu, et~al.
\newblock Highly scalable deep learning training system with mixed-precision:
  Training imagenet in four minutes.
\newblock \emph{arXiv preprint arXiv:1807.11205}, 2018.

\bibitem[Karimi et~al.(2019)Karimi, Derr, and Tang]{karimi2019characterizing}
Hamid Karimi, Tyler Derr, and Jiliang Tang.
\newblock Characterizing the decision boundary of deep neural networks.
\newblock \emph{arXiv preprint arXiv:1912.11460}, 2019.

\bibitem[Keskar et~al.(2017)Keskar, Mudigere, Nocedal, Smelyanskiy, and
  Tang]{KeskarMNST17}
Nitish~Shirish Keskar, Dheevatsa Mudigere, Jorge Nocedal, Mikhail Smelyanskiy,
  and Ping Tak~Peter Tang.
\newblock On large-batch training for deep learning: Generalization gap and
  sharp minima.
\newblock In \emph{International Conference on Learning Representations}, 2017.

\bibitem[Kumar et~al.(2021)Kumar, Wang, Young, Bradbury, Kumar, Chen, and
  Swing]{kumar2021exploring}
Sameer Kumar, Yu~Wang, Cliff Young, James Bradbury, Naveen Kumar, Dehao Chen,
  and Andy Swing.
\newblock Exploring the limits of concurrency in ml training on google tpus.
\newblock \emph{Machine Learning and Systems}, 2021.

\bibitem[Lee \& Raginsky(2017)Lee and Raginsky]{lee2017minimax}
Jaeho Lee and Maxim Raginsky.
\newblock Minimax statistical learning with wasserstein distances.
\newblock \emph{arXiv preprint arXiv:1705.07815}, 2017.

\bibitem[Li(2017)]{li2017scaling}
Mu~Li.
\newblock \emph{Scaling distributed machine learning with system and algorithm
  co-design}.
\newblock PhD thesis, PhD thesis, Intel, 2017.

\bibitem[Madry et~al.(2017)Madry, Makelov, Schmidt, Tsipras, and
  Vladu]{madry2017towards}
Aleksander Madry, Aleksandar Makelov, Ludwig Schmidt, Dimitris Tsipras, and
  Adrian Vladu.
\newblock Towards deep learning models resistant to adversarial attacks.
\newblock \emph{arXiv preprint arXiv:1706.06083}, 2017.

\bibitem[Martens \& Grosse(2015)Martens and Grosse]{martens2015optimizing}
James Martens and Roger Grosse.
\newblock Optimizing neural networks with kronecker-factored approximate
  curvature.
\newblock In \emph{International conference on machine learning}. PMLR, 2015.

\bibitem[Mattson et~al.(2019)Mattson, Cheng, Coleman, Diamos, Micikevicius,
  Patterson, Tang, Wei, Bailis, Bittorf, et~al.]{mattson2019mlperf}
Peter Mattson, Christine Cheng, Cody Coleman, Greg Diamos, Paulius
  Micikevicius, David Patterson, Hanlin Tang, Gu-Yeon Wei, Peter Bailis, Victor
  Bittorf, et~al.
\newblock Mlperf training benchmark.
\newblock \emph{arXiv preprint arXiv:1910.01500}, 2019.

\bibitem[Moosavi-Dezfooli et~al.(2019)Moosavi-Dezfooli, Fawzi, Uesato, and
  Frossard]{moosavi2019robustness}
Seyed-Mohsen Moosavi-Dezfooli, Alhussein Fawzi, Jonathan Uesato, and Pascal
  Frossard.
\newblock Robustness via curvature regularization, and vice versa.
\newblock In \emph{IEEE Conference on Computer Vision and Pattern Recognition},
  2019.

\bibitem[Osawa et~al.(2018)Osawa, Tsuji, Ueno, Naruse, Yokota, and
  Matsuoka]{osawa2018second}
Kazuki Osawa, Yohei Tsuji, Yuichiro Ueno, Akira Naruse, Rio Yokota, and Satoshi
  Matsuoka.
\newblock Second-order optimization method for large mini-batch: Training
  resnet-50 on imagenet in 35 epochs.
\newblock \emph{arXiv preprint arXiv:1811.12019}, 2018.

\bibitem[Papernot et~al.(2016)Papernot, McDaniel, and
  Goodfellow]{papernot2016transferability}
Nicolas Papernot, Patrick McDaniel, and Ian Goodfellow.
\newblock Transferability in machine learning: from phenomena to black-box
  attacks using adversarial samples.
\newblock \emph{arXiv preprint arXiv:1605.07277}, 2016.

\bibitem[Radford et~al.(2019)Radford, Wu, Child, Luan, Amodei, and
  Sutskever]{radford2019language}
Alec Radford, Jeffrey Wu, Rewon Child, David Luan, Dario Amodei, and Ilya
  Sutskever.
\newblock Language models are unsupervised multitask learners.
\newblock \emph{OpenAI blog}, 2019.

\bibitem[Shafahi et~al.(2019)Shafahi, Najibi, Ghiasi, Xu, Dickerson, Studer,
  Davis, Taylor, and Goldstein]{shafahi2019adversarial}
Ali Shafahi, Mahyar Najibi, Amin Ghiasi, Zheng Xu, John Dickerson, Christoph
  Studer, Larry~S Davis, Gavin Taylor, and Tom Goldstein.
\newblock Adversarial training for free!
\newblock \emph{arXiv preprint arXiv:1904.12843}, 2019.

\bibitem[Shallue et~al.(2018)Shallue, Lee, Antognini, Sohl-Dickstein, Frostig,
  and Dahl]{shallue2018measuring}
Christopher~J Shallue, Jaehoon Lee, Joseph Antognini, Jascha Sohl-Dickstein,
  Roy Frostig, and George~E Dahl.
\newblock Measuring the effects of data parallelism on neural network training.
\newblock \emph{arXiv preprint arXiv:1811.03600}, 2018.

\bibitem[Sinha et~al.(2017)Sinha, Namkoong, and Duchi]{sinha2017certifiable}
Aman Sinha, Hongseok Namkoong, and John Duchi.
\newblock Certifiable distributional robustness with principled adversarial
  training.
\newblock \emph{arXiv preprint arXiv:1710.10571}, 2017.

\bibitem[Smith et~al.(2017)Smith, Kindermans, Ying, and Le]{smith2017don}
Samuel~L Smith, Pieter-Jan Kindermans, Chris Ying, and Quoc~V Le.
\newblock Don't decay the learning rate, increase the batch size.
\newblock \emph{arXiv preprint arXiv:1711.00489}, 2017.

\bibitem[Wang et~al.(2019)Wang, Ma, Bailey, Yi, Zhou, and
  Gu]{wang2019convergence}
Yisen Wang, Xingjun Ma, James Bailey, Jinfeng Yi, Bowen Zhou, and Quanquan Gu.
\newblock On the convergence and robustness of adversarial training.
\newblock In \emph{ICML}, volume~1, pp.\ ~2, 2019.

\bibitem[Wong et~al.(2020)Wong, Rice, and Kolter]{wong2020fast}
Eric Wong, Leslie Rice, and J~Zico Kolter.
\newblock Fast is better than free: Revisiting adversarial training.
\newblock \emph{arXiv preprint arXiv:2001.03994}, 2020.

\bibitem[Xie et~al.(2020)Xie, Tan, Gong, Wang, Yuille, and
  Le]{xie2020adversarial}
Cihang Xie, Mingxing Tan, Boqing Gong, Jiang Wang, Alan~L Yuille, and Quoc~V
  Le.
\newblock Adversarial examples improve image recognition.
\newblock In \emph{IEEE Conference on Computer Vision and Pattern Recognition},
  2020.

\bibitem[Yamazaki et~al.(2019)Yamazaki, Kasagi, Tabuchi, Honda, Miwa, Fukumoto,
  Tabaru, Ike, and Nakashima]{yamazaki2019yet}
Masafumi Yamazaki, Akihiko Kasagi, Akihiro Tabuchi, Takumi Honda, Masahiro
  Miwa, Naoto Fukumoto, Tsuguchika Tabaru, Atsushi Ike, and Kohta Nakashima.
\newblock Yet another accelerated sgd: Resnet-50 training on imagenet in 74.7
  seconds.
\newblock \emph{arXiv preprint arXiv:1903.12650}, 2019.

\bibitem[Yao et~al.(2018{\natexlab{a}})Yao, Gholami, Arfeen, Liaw, Gonzalez,
  Keutzer, and Mahoney]{yao2018large}
Zhewei Yao, Amir Gholami, Daiyaan Arfeen, Richard Liaw, Joseph Gonzalez, Kurt
  Keutzer, and Michael Mahoney.
\newblock Large batch size training of neural networks with adversarial
  training and second-order information.
\newblock \emph{arXiv preprint arXiv:1810.01021}, 2018{\natexlab{a}}.

\bibitem[Yao et~al.(2018{\natexlab{b}})Yao, Gholami, Lei, Keutzer, and
  Mahoney]{yao2018hessian}
Zhewei Yao, Amir Gholami, Qi~Lei, Kurt Keutzer, and Michael~W Mahoney.
\newblock Hessian-based analysis of large batch training and robustness to
  adversaries.
\newblock \emph{arXiv preprint arXiv:1802.08241}, 2018{\natexlab{b}}.

\bibitem[Ying et~al.(2018)Ying, Kumar, Chen, Wang, and Cheng]{ying2018image}
Chris Ying, Sameer Kumar, Dehao Chen, Tao Wang, and Youlong Cheng.
\newblock Image classification at supercomputer scale.
\newblock \emph{arXiv preprint arXiv:1811.06992}, 2018.

\bibitem[You et~al.(2017)You, Gitman, and Ginsburg]{you2017scaling}
Yang You, Igor Gitman, and Boris Ginsburg.
\newblock Scaling sgd batch size to 32k for imagenet training.
\newblock \emph{arXiv preprint arXiv:1708.03888}, 2017.

\bibitem[You et~al.(2018)You, Zhang, Hsieh, Demmel, and
  Keutzer]{you2018imagenet}
Yang You, Zhao Zhang, Cho-Jui Hsieh, James Demmel, and Kurt Keutzer.
\newblock Imagenet training in minutes.
\newblock In \emph{International Conference on Parallel Processing}, 2018.

\bibitem[You et~al.(2019)You, Hseu, Ying, Demmel, Keutzer, and
  Hsieh]{you2019large}
Yang You, Jonathan Hseu, Chris Ying, James Demmel, Kurt Keutzer, and Cho-Jui
  Hsieh.
\newblock Large-batch training for lstm and beyond.
\newblock In \emph{Proceedings of the International Conference for High
  Performance Computing, Networking, Storage and Analysis}, 2019.

\bibitem[Zhang et~al.(2019)Zhang, Zhang, Lu, Zhu, and Dong]{zhang2019you}
Dinghuai Zhang, Tianyuan Zhang, Yiping Lu, Zhanxing Zhu, and Bin Dong.
\newblock You only propagate once: Accelerating adversarial training via
  maximal principle.
\newblock \emph{arXiv preprint arXiv:1905.00877}, 2019.

\end{thebibliography}
\bibliographystyle{iclr2022_conference}

\newpage

\appendix
\section{Appendix}
\subsection{The proof of Lemma 1:}
\label{Appendix:1}

Under Assumptions 1 and 2, we have \\

\begin{equation}
\|x_{i}^{*}(\theta_{t}) - x_{i}^{*}(\theta_{t-\tau})\| \leq \frac{L_{x \theta}}{\mu}\|\theta_{t} - \theta_{t-\tau}\| \leq \frac{L_{x \theta}}{\mu} \eta \tau M 
\end{equation}

where  $x_{i}^{*}(\theta_{t})$ and $x_{i}^{*}(\theta_{t-\tau})$ denote the adversarial example of $x_{i}$ calculated by $\theta_{t}$ and stale weight $\theta_{t-\tau}$, respectively.

Proof: 

 According to Assumption \ref{assumption_2}, we have

\begin{equation}
\label{1-1}
\begin{split}
\mathcal{L}(\theta_{t}, x_{i}^{*}(\theta_{t-\tau})) &\leq \mathcal{L}(\theta_{t}, x_{i}^{*}(\theta_{t})) +
\langle \nabla_{x}\mathcal{L}(\theta_{t}, x_{i}^{*}(\theta_{t})), x_{i}^{*}(\theta_{t-\tau}) - x_{i}^{*}(\theta_{t}) \rangle - \\
&\frac{\mu}{2} \| x_{i}^{*}(\theta_{t - \tau}) - x_{i}^{*}(\theta_{t}) \|^{2}, \\
&\leq \mathcal{L}(\theta_{t}, x_{i}^{*}(\theta_{t})) - \frac{\mu}{2} \|x_{i}^{*}(\theta_{t-\tau}) - x_{i}^{*}(\theta_{t})\|^{2}
\end{split}
\end{equation}

In addition, we have

\begin{equation}
\label{1-2}
\begin{split}
    \mathcal{L}(\theta_{t}, x_{i}^{*}(\theta_{t})) &\leq \mathcal{L}(\theta_{t}, x_{i}^{*}(\theta_{t-\tau})) + \langle \nabla_{x}\mathcal{L}(\theta_{t}, x_{i}^{*}(\theta_{t-\tau})), x_{i}^{*}(\theta_{t}) - x_{i}^{*}(\theta_{t-\tau})\rangle - \\ & \frac{\mu}{2}\|x_{i}^{*}(\theta_{t-\tau}) - x_{i}^{*}(\theta_{t})\|^{2}
\end{split}
\end{equation}

Combining (\ref{1-1}) and (\ref{1-2}), we can obtain:

\begin{equation}
\begin{split}
    \mu \|x_{i}^{*}(\theta_{t-\tau}) - x_{i}^{*}(\theta_{t})\|^{2} &\leq \langle \nabla_{x} \mathcal{L}(\theta_{t}, x_{i}^{*}(\theta_{t})), x_{i}^{*}(\theta_{t}) - x_{i}^{*}(\theta_{t-\tau}) \rangle \\
    &\leq \langle\nabla_{x}\mathcal{L}(\theta_{t}, x_{i}^{*}(\theta_{t-\tau})) - \nabla_{x}\mathcal{L}(\theta_{t-\tau}, x_{i}^{*}(\theta_{t-\tau})), x_{i}^{*}(\theta_{t}) - x_{i}^{*}(\theta_{t-\tau})\rangle \\
    &\leq \|\nabla_{x} \mathcal{L}(\theta_{t}, x_{i}^{*}(\theta_{t-\tau})) - \nabla_{x}\mathcal{L}(\theta_{t-\tau}, x_{i}^{*}(\theta_{t-\tau}))\| \|x_{i}^{*}(\theta_{t}) - x_{i}^{*}(\theta_{t-\tau})\|\\
    &\leq L_{x \theta}\|\theta_{t} - \theta_{t-\tau}\| \|x_{i}^{*}(\theta_{t}) - x_{i}^{*}(\theta_{t-\tau})\|
\end{split}
\end{equation}

where the second inequality is due to $\langle \nabla_{x}\mathcal{L}(\theta_{t-\tau}, x_{i}^{*}(\theta_{t-\tau})), x_{i}^{*}(\theta_{t}) - x_{i}^{*}(\theta_{t-\tau})\rangle \leq 0$, the third inequality holds because CauchySchwarz inequality and the last inequality follows from Assumption \ref{assumption_1}.
Therefore, 

\begin{equation}
\begin{split}
    \|x_{i}^{*}(\theta_{t}) - x_{i}^{*}(\theta_{t-\tau})\| &\leq \frac{L_{x \theta}}{\mu}\|\theta_{t} - \theta_{t-\tau}\| \\
    &\leq \frac{L_{x \theta}}{\mu}\|\sum_{j\in[1,\tau]}(\theta_{t-j+1} - \theta_{t-j})\| \\
    &\leq \frac{L_{x \theta}}{\mu} \| \sum_{j \in [1,\tau]} \eta \hat{g}_{t-j}(x_{i}))  \| \\
    &\leq \frac{L_{x \theta}}{\mu} \eta \tau M
\end{split}
\end{equation}

where the second inequality follows the calculation of delayed weight, the third inequality holds because the difference of weights is calculated with gradient $\hat{g}_{t-j} (j\in[1,\tau])$ and the last inequality holds follows 
Assumption 3.

Thus, 

\begin{equation}
    \|x_{i}^{*}(\theta_{t}) - x_{i}^{*}(\theta_{t-\tau})\| \leq \frac{L_{x\theta}}{\mu} \eta\tau M
\end{equation}

This completes the proof.

\subsection{}
\label{Appendix:2}
\begin{lemma}

Under Assumptions 1 and 2, we have $\mathcal{L}_{D}(\theta)$ is L-smooth where $L=L_{\theta \theta} +  \frac{L_{x \theta}}{2\mu}L_{\theta x}$, i.e., for any $\theta_{1}$ and $\theta_{2}$, we can say
\begin{equation}
\begin{split}
\|\nabla_{\theta} \mathcal{L}_{D}(\theta_{t}) - \nabla_{\theta} \mathcal{L}_{D}(\theta_{t-\tau})\| \leq L\|\theta_{t} - \theta_{t-\tau}\|
\end{split}
\end{equation}
\begin{equation}
\begin{split}
\mathcal{L}_{D}(\theta_{t}) = \mathcal{L}_{D}(\theta_{t-\tau}) + \langle \mathcal{L}_{D}(\theta_{t-\tau}), \theta_{t} - \theta_{t-\tau} \rangle + \frac{L}{t-\tau} \|\theta_{t} - \theta_{t-\tau} \|
\end{split}
\end{equation}
\end{lemma}

Proof:

Based on Lemma 1, we can obtain:

\begin{equation}
    \|x_{i}^{*}(\theta_{t}) - x_{i}^{*}(\theta_{t-\tau})\| \leq \frac{L_{x \theta}}{\mu} \|\theta_{t} - \theta_{t-\tau}\|
\end{equation}

We can obtain for $i \in [n]$:

\begin{equation}
\begin{split}
    \|\nabla_{\theta} \mathcal{L}(\theta_{t}, x_{i}^{*}(\theta_{t})) &- \nabla_{\theta} \mathcal{L}(\theta_{t-\tau}, x_{i}^{*}(\theta_{t-\tau}))\| \\ 
    \leq &\|\nabla_{\theta} \mathcal{L}(\theta_{t}, x_{i}^{*}(\theta_{t})) - \nabla_{\theta} \mathcal{L}(\theta_{t}, x_{i}^{*}(\theta_{t-\tau}))\| \\ 
    &+ \|\nabla_{\theta} \mathcal{L}(\theta_{t}, x_{i}^{*}(\theta_{t-\tau})) - \nabla_{\theta} \mathcal{L}(\theta_{t-\tau}, x_{i}^{*}(\theta_{t-\tau}))\| \\
    \leq &L_{\theta \theta} \|\theta_{t} - \theta_{t-\tau}\|  +     L_{\theta x} \|x_{i}^{*}(\theta_{t}) - x_{i}^{*}(\theta_{t-\tau})\|\\
    \leq &L_{\theta \theta} \|\theta_{t} - \theta_{t-\tau} \| + L_{\theta x}\frac{L_{x\theta}}{\mu}\|\theta_{t} - \theta_{t-\tau}\| \\
    = &(L_{\theta \theta} + L_{\theta x}\frac{L_{x \theta}}{\mu}) \|\theta_{t} - \theta_{t-\tau} \| \\
\end{split}
\end{equation}

where the second inequality holds because Assumption 1, the third inequality holds follows Lemma 1.

Therefore,

\begin{equation}
\begin{split}
    \|\nabla \mathcal{L}_{D}(\theta_{t}) - \nabla \mathcal{L}_{D}(\theta_{t-\tau})\rVert_{2} &\leq \|\frac{1}{2n}\sum_{i=1}^{n} (\nabla_{\theta} \mathcal{L}(\theta_{t}, x_{i}) + \nabla_{\theta} \mathcal{L}(\theta_{t}, x_{i}^{*}(\theta_{t}))) \\
    & \quad - \frac{1}{2n} \sum_{i=1}^{n} (\nabla_{\theta} \mathcal{L}(\theta_{t-\tau}, x_{i}) + \nabla_{\theta} \mathcal{L}(\theta_{t-\tau}, x_{i}^{*}(\theta_{t-\tau})))\| \\
    &\leq \frac{1}{2n}\sum_{i=1}^{n}\|\nabla_{\theta} \mathcal{L}(\theta_{t}, x_{i}) - \nabla_{\theta} \mathcal{L}(\theta_{t-\tau}, x_{i})\| \\
    & \quad + \frac{1}{2n} \sum_{i=1}^{n} \|\nabla_{\theta} \mathcal{L}(\theta_{t}, x_{i}^{*}(\theta_{t})) - \nabla_{\theta} \mathcal{L}(\theta_{t-\tau}, x_{i}^{*}(\theta_{t-\tau}))\| \\
    &\leq \frac{1}{2}L_{\theta \theta}\|\theta_{t} - \theta_{t-\tau}\| + \frac{1}{2} (L_{\theta \theta} + L_{\theta x} \frac{L_{x \theta}}{\mu}) \|\theta_{t} - \theta_{t-\tau}\| \\
    &= (L_{\theta \theta} +  \frac{L_{x \theta}}{2\mu}L_{\theta x}) \|\theta_{t} - \theta_{t-\tau}\|
\end{split}
\end{equation}

This completes the proof.

\subsection*{A.3}
\label{Appendix:3}
\begin{lemma}

Let $\hat{x}_{i}(\theta_{t})$ be the $\lambda$-solution of $x_{i}^{*}(\theta_{t})$: $  \langle   x_{i}^{*}(\theta_{t}) - \hat{x}_{i}(\theta_{t}), \nabla_{x}\mathcal{L}(\theta_{t}, \hat{x}_{i}(\theta_{t})) \rangle \leq \lambda $. Under Assumptions 1 and 2, for the concurrent stochastic gradient $\hat{g}(\theta)$, we have

\begin{equation}
    \|g(\theta_{t}) - \hat{g}(\theta_{t-\tau})\| \leq \frac{L_{\theta x}}{2}\sqrt{\frac{\lambda}{\mu}}.
\end{equation}

\end{lemma}
Proof:

\begin{equation}
\begin{split}
    \|g(\theta_{t}) - \hat{g}(\theta_{t-\tau})\| 
    &= \|\frac{1}{2|\mathcal{B}|}\sum_{i \in \mathcal{|B|}}(\nabla_{\theta}\mathcal{L}(\theta_{t}, x_{i}) + \nabla_{\theta}\mathcal{L}(\theta_{t}, x_{i}^{*}(\theta_{t})) \\ 
    &\quad - (\nabla_{\theta}\mathcal{L}(\theta_{t}, x_{i}) + \nabla_{\theta}\mathcal{L}(\theta_{t}, \hat{x}_{i}(\theta_{t-\tau})))\| \\
    &= \|\frac{1}{2|\mathcal{B}|}\sum_{i \in \mathcal{|B|}}(\nabla_{\theta}\mathcal{L}(\theta_{t}, x_{i}^{*}(\theta_{t})) - \nabla_{\theta}\mathcal{L}(\theta_{t}, \hat{x}_{i}(\theta_{t-\tau})))\| \\
    &\leq \frac{1}{2|\mathcal{B}|}\sum_{i \in \mathcal{|B|}}\|\nabla_{\theta}\mathcal{L}(\theta_{t}, x_{i}^{*}(\theta_{t})) - \nabla_{\theta}\mathcal{L}(\theta_{t}, \hat{x}_{i}(\theta_{t-\tau}))\| \\
    &\leq \frac{1}{2|\mathcal{B}|}\sum_{i\in\mathcal{|B|}}L_{\theta x}\|x_{i}^{*}(\theta_{t}) - \hat{x}_{i}(\theta_{t-\tau})\| \\
    &= \frac{1}{2|\mathcal{B}|}\sum_{i\in\mathcal{|B|}}L_{\theta x}\|x_{i}^{*}(\theta_{t}) - x_{i}^{*}(\theta_{t-\tau}) + x_{i}^{*}(\theta_{t-\tau}) - \hat{x}_{i}(\theta_{t-\tau})\| \\
    &\leq \frac{1}{2|\mathcal{B}|}\sum_{i\in\mathcal{|B|}}(L_{\theta x}\|x_{i}^{*}(\theta_{t}) - x_{i}^{*}(\theta_{t-\tau})\| + L_{\theta x}\|x_{i}^{*}(\theta_{t-\tau}) - \hat{x}_{i}(\theta_{t-\tau})\|) \\
    &= \frac{1}{2|\mathcal{B}|}\sum_{i\in\mathcal{|B|}}(L_{\theta x}\|x_{i}^{*}(\theta_{t}) - x_{i}^{*}(\theta_{t-\tau})\| + L_{\theta x}\|x_{i}^{*}(\theta_{t-\tau}) - \hat{x}_{i}(\theta_{t-\tau})\|) \\
    &\leq \frac{1}{2|\mathcal{B}|}\sum_{i\in\mathcal{|B|}}(L_{\theta x}  \eta \tau M \frac{L_{x \theta}}{\mu}   + L_{\theta x}\|x_{i}^{*}(\theta_{t-\tau}) - \hat{x}_{i}(\theta_{t-\tau})\|)
\end{split}
\end{equation}

Let $\hat{x}_{i}(\theta_{t-\tau})$ be the $\lambda$-approximate of $x_{i}^{*}(\theta_{t-\tau})$, we can obtain:
\begin{equation}
\label{approximate_1}
    \langle x_{i}^{*}(\theta_{t-\tau}) - \hat{x}_{i}(\theta_{t-\tau}), \nabla_{\theta}\mathcal{L}(\theta_{t-\tau}; \hat{x}_{i}(\theta_{t-\tau})) \rangle \leq \delta
\end{equation}

In addition, we can obtain:

\begin{equation}
\label{approximate_2}
    \langle \hat{x}_{i}(\theta_{t-\tau}) - x_{i}^{*}(\theta_{t-\tau}), \nabla_{x}\mathcal{L}(\theta_{t-\tau}, x_{i}^{*}(\theta_{t-\tau})) \leq 0
\end{equation}

Combining \ref{approximate_1} and \ref{approximate_2}, we have: 

\begin{equation}
\label{approximate_7}
    \langle x_{i}^{*}(\theta_{t-\tau}) - \hat{x}_{i}(\theta_{t-\tau}), \nabla_{\theta}\mathcal{L}(\theta_{t-\tau}; \hat{x}_{i}(\theta_{t-\tau})) - \nabla_{x}\mathcal{L}(\theta_{t-\tau}, x_{i}^{*}(\theta_{t-\tau})) \rangle \leq \lambda
\end{equation}

Based on Assumption 2, we have 

\begin{equation}
\label{approximate_3}
    \mu \|x_{i}^{*}(\theta_{t-\tau}) - \hat{x}_{i}(\theta_{t-\tau})\|^{2} \leq \langle \nabla_{x} \mathcal{L}(\theta_{t-\tau}, x_{i}^{*}(\theta_{t-\tau})) - \nabla_{x} \mathcal{L}(\theta_{t-\tau}, \hat{x}_{i}(\theta_{t-\tau}), \hat{x}_{i} - x_{i}^{*}(\theta_{t-\tau})) \rangle 
\end{equation}

Combining \ref{approximate_3} with \ref{approximate_7}, we can obtain:

\begin{equation}
    \mu \|x_{i}^{*}(\theta_{t-\tau}) - \hat{x}(\theta_{t-\tau})\|^{2} \leq \lambda
\end{equation}

Therefore, we have

\begin{equation}
    \|x_{i}^{*}(\theta_{t-\tau}) - \hat{x}(\theta_{t-\tau})\| \leq \sqrt{\frac{\lambda}{\mu}}.
\end{equation}

Thus, we can obtain
\begin{equation}
    \|g(\theta_{t}) - \hat{g}(\theta_{t-\tau})\| \leq  \frac{L_{\theta x}}{2}( \eta \tau M \frac{L_{x \theta}}{\mu} + \sqrt{\frac{\lambda}{\mu}})
\end{equation}

This completes the proof.

\subsection{The proof of Theorem 1:}
\label{Appendix:4}

Suppose Assumptions 1, 2, 3 and 4 hold. Let $\Delta = \mathcal{L}_{D}(\theta_{0}) - \min_{\theta} \mathcal{L}_{D}(\theta)$, $\nabla\mathcal{L}_{D}(\theta_{t}) =\frac{1}{2n}\sum_{i=1}^{n}( \nabla\mathcal{L}(\theta_{t}, x_{i}, y_{i}) + \nabla\mathcal{L}(\theta_{t}, x_{i}^{*}, y_{i}))$.   If the step size of outer minimization is set to $\eta_{t} = \eta = \min(1/L, \sqrt{\Delta / L \sigma^{2}T})$, where $L=L_{\theta \theta} +  \frac{L_{x \theta}}{2\mu}L_{\theta x}$. Then the output of Algorithm 1 satisfies:

\begin{equation}
        \frac{1}{T}\sum_{t=0}^{T-1}\mathbb{E}[\|\nabla \mathcal{L}_{D}(\theta_{t})\|^{2}] \leq 2\sigma \sqrt{\frac{L\Delta}{T}} + \frac{L_{\theta x}^{2} }{2}(\frac{\tau M L_{x \theta}}{L \mu} + \sqrt{\frac{\lambda}{\mu}})^{2}
\end{equation}
where $L=(ML_{\theta x}L_{x \theta}/\epsilon\mu + L_{\theta \theta})$.

Proof:

\begin{equation}
\label{3-1}
\begin{split}
    \mathcal{L}_{D}(\theta_{t+1}) &\leq \mathcal{L}_{D}(\theta_{t}) + \langle \nabla\mathcal{L}_{D}(\theta_{t}), \theta_{t+1} - \theta_{t}\rangle + \frac{L}{2}\|\theta_{t+1} - \theta_{t}\|^{2} \\
    &= \mathcal{L}_{D}(\theta_{t}) - \eta \|\nabla \mathcal{L}_{D}(\theta_{t}) \|^{2} + \frac{L\eta^{2}}{2}||\hat{g}(\theta_{t})\|_{2}^{2} + \eta \langle \nabla \mathcal{L}_{D}(\theta_{t}), \nabla \mathcal{L}_{D}(\theta_{t}) - \hat{g}(\theta_{t}) \rangle \\
    &= \mathcal{L}_{D}(\theta_{t}) - \eta(1 - \frac{L\eta}{2}) \|\nabla \mathcal{L}_{D}(\theta_{t})\|^{2} + \eta(1-L\eta) \langle \nabla \mathcal{L}_{D}(\theta_{t}), \nabla \mathcal{L}_{D}(\theta_{t}) - \hat{g}(\theta_{t}) \rangle \\
    &\quad + \frac{L\eta^{2}}{2} \|\hat{g}(\theta_{t}) - \nabla \mathcal{L}_{D}(\theta_{t}) \|^{2} \\
    &= \mathcal{L}_{D}(\theta_{t}) - \eta(1-\frac{L\eta}{2})\|\nabla \mathcal{L}_{D}(\theta_{t})\|^{2} + \eta(1-L\eta) \langle \nabla\mathcal{L}_{D}(\theta_{t}), g(\theta_{t}) - \hat{g}(\theta_{t}) \rangle \\
    &+ \eta(1-L\eta)\langle \nabla \mathcal{L}_{D}(\theta_{t}), \nabla \mathcal{L}_{D}(\theta_{t}) - g(\theta_{t}) \rangle + \frac{L\eta^{2}}{2}\|\hat{g}(\theta_{t}) - g(\theta_{t}) + g(\theta_{t}) - \nabla \mathcal{L}_{D}(\theta_{t})\|^{2} \\
    &\leq \mathcal{L}_{D}(\theta_{t}) - \frac{\eta}{2} \lVert \nabla \mathcal{L}_{D}(\theta_{t})\|_{2}^{2} + \frac{\eta}{2}(1-L\eta)\|\hat{g}(\theta_{t}) - g(\theta_{t})\|^{2} \\
    &\quad + \eta (1-L\eta) \langle \nabla\mathcal{L}_{D}(\theta_{t}), \nabla \mathcal{L}_{D}(\theta_{t}) - g(\theta_{t}) \rangle + L\eta^{2}(\|\hat{g}(\theta_{t}) - g(\theta) \|_{2}^{2} + \lVert g(\theta_{t}) - \nabla \mathcal{L}_{D}(\theta_{t})\|^{2}) \\
    &= \mathcal{L}_{D}(\theta_{t}) - \frac{\eta}{2}||\nabla \mathcal{L}_{D}(\theta_{t})\|_{2}^{2} 
    + \frac{\eta}{2}(1+L \eta) \|\hat{g}(\theta_{t}) - g(\theta_{t})\|^{2} \\
    &\quad + \eta(1-L\eta)\langle \nabla \mathcal{L}_{D}(\theta_{t}), \nabla \mathcal{L}_{D}(\theta_{t}) - g(\theta_{t})\rangle  + L\eta^{2}\|g(\theta_{t}) - \nabla \mathcal{L}_{D}(\theta_{t})\|_{2}^{2})
\end{split}
\end{equation}

Taking expectation on both sides of the above inequality conditioned on $\theta_{t}$, we can obtain:

\begin{equation}
\label{3-2}
\begin{split}
    \mathbb{E}[\mathcal{L}_{D}(\theta_{t+1}) - \mathcal{L}_{D}(\theta_{t})|\theta_{t}] &\leq -\frac{\eta}{2}||\nabla\mathcal{L}_{D}(\theta_{t})\|^{2} + \frac{\eta}{2}(1+L\eta)(\frac{L_{\theta x}}{2}(\eta \tau M \frac{L_{x \theta}}{\mu} + \sqrt{\frac{\lambda}{\mu}}))^{2}  + L\eta^{2}\sigma^{2} \\
    &= -\frac{\eta}{2}\|\nabla\mathcal{L}_{D}(\theta_{t})\|^{2} + \frac{\eta}{8}(1+L\eta)(L_{\theta x}(\eta \tau M \frac{L_{x \theta}}{\mu} + \sqrt{\frac{\lambda}{\mu}}))^{2}  + L\eta^{2}\sigma^{2} \\
    &= -\frac{\eta}{2}\|\nabla\mathcal{L}_{D}(\theta_{t})\|^{2} + \frac{\eta L_{\theta x}^{2}}{8}(1+L\eta)(\eta \tau M \frac{L_{x \theta}}{\mu} +  \sqrt{\frac{ \lambda}{\mu}})^{2}  + L\eta^{2}\sigma^{2} \\
\end{split}
\end{equation}

where we used the fact that $\mathbb{E}[g(\theta_{t})]=\nabla \mathcal{L}_{D}(\theta_{t})$, Assumption 2, Lemma 2 and Lemma 3. Taking the sum of (\ref{3-2}) over $t=0, ..., T-1$, we obtain that:

\begin{equation}
\label{3-3}
    \sum_{t=0}^{T-1}\frac{\eta}{2}\mathbb{E}[\|\nabla\mathcal{L}_{D}(\theta_{t})\|^{2}] \leq \mathbb{E}[\mathcal{L}_{D}(\theta_{0}) - \mathcal{L}_{D}(\theta_{T})] + \sum_{t=0}^{T-1}\frac{\eta L_{\theta x}^{2}}{8}(1+L\eta)(\eta \tau M \frac{L_{x \theta}}{\mu} +  \sqrt{\frac{ \lambda}{\mu}})^{2} + L\sum_{t=0}^{T-1}\eta^{2}\sigma^{2}
\end{equation}

Choose $\eta = \min(1/L, \sqrt{\frac{\Delta}{TL\sigma^{2}}})$ where $\Delta = \mathcal{L}_{D}(\theta_{0}) - \mathcal{L}_{D}(\theta_{T})$ and 
$L=L_{\theta \theta} +  \frac{L_{x \theta}}{2\mu}L_{\theta x}$, we can show that:

\begin{equation}
    \frac{1}{T}\sum_{t=0}^{T-1}\mathbb{E}[\|\nabla \mathcal{L}_{D}(\theta_{t})\|^{2}] \leq 2\sigma \sqrt{\frac{L\Delta}{T}} + \frac{L_{\theta x}^{2} }{2}(\frac{\tau M L_{x \theta}}{L \mu} + \sqrt{\frac{\lambda}{\mu}})^{2}
\end{equation}

% This completes the proof. The proof is inspired by \cite{sinha2017certifiable,wang2019convergence}.

\subsection{Hyperparameters}
\label{Appendix:5}

\subsubsection*{Hyperparameters:}

More specially, our main hyperparameters are shown in Table \ref{resnet_hyperparam}.

\begin{table}[H]
\renewcommand\arraystretch{1.2}
\caption{Hyperparameters of ResNet-50 on ImageNet}
\label{resnet_hyperparam}
\vskip 0.15in
\begin{center}
\begin{sc}
\begin{tabular}{lccc}

\ & \multicolumn{1}{c}{\bf 32k} & \multicolumn{1}{c}{\bf 64k} & \multicolumn{1}{c}{\bf 96k} \\
\hline
\multicolumn{1}{l}{\bf Peak LR}    & 35.0  & 41.0 & 43.0  \\
\multicolumn{1}{l}{\bf Epoch}    & 90  & 90 & 90  \\
\multicolumn{1}{l}{\bf Weight Decay} & 5e-4  & 5e-4 & 5e-4 \\
\multicolumn{1}{l}{\bf Warmup} & 40   & 41   & 41 \\
\multicolumn{1}{l}{\bf LR decay} & poly & poly & poly \\
\multicolumn{1}{l}{\bf Optimizer} & LARS & LARS & LARS \\
\multicolumn{1}{l}{\bf Momentum} & 0.9 & 0.9 & 0.9\\
\multicolumn{1}{l}{\bf Label Smoothing} & 0.1 & 0.1 & 0.1 \\
\hline
\end{tabular}
\end{sc}
\end{center}
\vskip -0.1in
\end{table}

\end{document}